\documentclass[10pt,journal,compsoc]{IEEEtran}
\usepackage{algorithmic}
\usepackage{algorithm}
\usepackage{array}
\usepackage[caption=false,font=normalsize,labelfont=sf,textfont=sf]{subfig}
\usepackage{textcomp}
\usepackage{stfloats}
\usepackage{url}
\usepackage{cite}
\usepackage{multirow}
\newcommand{\myparagraph}[1]{\vspace{0.1em}\noindent\textbf{#1}}
\usepackage[colorlinks,hyperindex,breaklinks]{hyperref}
\newcommand{\ie}{\textit{i}.\textit{e}.}
\newcommand{\eg}{\textit{e}.\textit{g}.}

\usepackage{pifont}

\usepackage{color}
\usepackage{graphicx}
\usepackage{ragged2e}
\usepackage{arydshln}
\usepackage{ulem}
\begin{document}
\title{Understanding the Tricks of Deep Learning in Medical Image Segmentation: Challenges and Future Directions}
\author{Dong~Zhang$^\dag$,
        Yi~Lin$^\dag$,
        Hao~Chen$^*$,
        Zhuotao~Tian,
        Xin~Yang,
        Jinhui~Tang,
        Kwang-Ting~Cheng,~\IEEEmembership{Fellow,~IEEE}
\IEEEcompsocitemizethanks{
\IEEEcompsocthanksitem D. Zhang, Y. Lin, H. Chen, and K. Cheng are with the Department of Computer Science and Engineering, The Hong Kong University of Science and Technology, Hong Kong, China. E-mail:~\{dongz, jhc, timcheng\}@ust.hk; yi.lin@connect.ust.hk.
\IEEEcompsocthanksitem Z. Tian is with SmartMore Inc, China. E-mail: zttian@cse.cuhk.edu.hk.
\IEEEcompsocthanksitem X. Yang is with the Department of Electronic Information and Communications, Huazhong University of Science and Technology, Wuhan 430074, China. E-mail: xinyang2014@hust.edu.cn.
\IEEEcompsocthanksitem J. Tang is with the School of Computer Science and Engineering, Nanjing University of Science and Technology, Nanjing 210094, China. E-mail: jinhuitang@njust.edu.cn.
}
\thanks{$\dag$~These two authors contributed equally to this work.}
\thanks{$*$~Corresponding author: Hao Chen.}
}
\markboth{Under Submission}%
{Shell \MakeLowercase{\textit{et al.}}: Bare Demo of IEEEtran.cls for Computer Society Journals}
\IEEEtitleabstractindextext{
\begin{abstract}
\justifying
Over the past few years, the rapid development of deep learning technologies for computer vision has significantly improved the performance of medical image segmentation (MedISeg). However, the diverse implementation strategies of various models have led to an extremely complex MedISeg system, resulting in a potential problem of unfair result comparisons. In this paper, we collect a series of MedISeg tricks for different model implementation phases (\ie, pre-training model, data pre-processing, data augmentation, model implementation, model inference, and result post-processing), and experimentally explore the effectiveness of these tricks on consistent baselines. With the extensive experimental results on both the representative 2D and 3D medical image datasets, we explicitly clarify the effect of these tricks. Moreover, based on the surveyed tricks, we also open-sourced a strong MedISeg repository, where each component has the advantage of plug-and-play. We believe that this milestone work not only completes a comprehensive and complementary survey of the state-of-the-art MedISeg approaches, but also offers a practical guide for addressing the future medical image processing challenges including but not limited to small dataset, class imbalance learning, multi-modality learning, and domain adaptation. The code and training weights have been released at:~\href{https://github.com/hust-linyi/seg_trick}{MedISeg}.
\end{abstract}
}

\maketitle
\IEEEdisplaynontitleabstractindextext
\IEEEpeerreviewmaketitle
\IEEEraisesectionheading{\section{Introduction}
\label{sec:introduction}}
\IEEEPARstart{M}{edical} image segmentation (MedISeg) is one of the most representative and comprehensive research topics in both communities of computer vision and medical image analysis~\cite{lynch2018new,wang2022boundary,azad2022medical}. It can not only recognize the object category but also locate the pixel-level positions~\cite{wang2018deepigeos,shen2017deep,zhu2021dual,liu2016relationship,li2018h,liu2022deep}. In clinical practice, MedISeg has been successfully used 
in a wide range of potential applications with qualitative and quantitative analyses, \eg, cancer diagnosis~\cite{munir2019cancer}, tumor change detection~\cite{jin2021predicting}, treatment planning~\cite{pham2000current}, and computer-integrated surgery~\cite{taylor2020computer}. To achieve satisfactory performance, one of the challenges is to enable the segmentation model to learn a set of rich yet discriminative features effectively and efficiently~\cite{fu2019dual,zhang2018recursive,lu2019see,long2015fully}.

\begin{figure}[t]
\centering
\includegraphics[width=.4\textwidth]{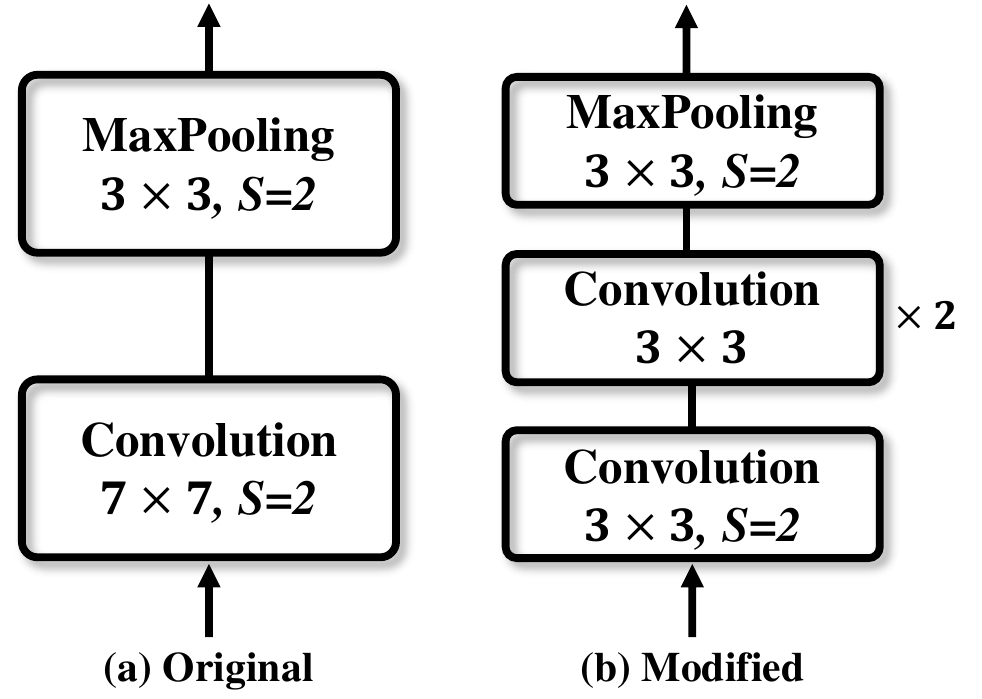}
\vspace{-2mm}
\caption{Two implementation schemes of the input stem in ResNet~\cite{he2016deep}, where (a) is the original implementation as claimed in its paper, and (b) is the modified implementation to reduce the computational costs. ``\emph{S}'' denotes the stride size. ``$\times 2$'' denotes that this block is repeated twice.}
\vspace{-2mm}
\label{fig1}
\end{figure} 
In recent years, the performance of MedISeg has been greatly improved~\cite{chen2018voxresnet,lin2023rethinking,shen2017deep,yang2017co,zhang2011multimodal,zhang2015deep,suk2014hierarchical,zhang2012multi,shan2020lung,ge2022x}, thanks to the remarkable progress in deep image processing technologies~\cite{long2015fully,chen2017rethinking,szegedy2015going,huang2017densely,he2016deep}. 
Advanced backbone networks (\eg, AlexNet~\cite{krizhevsky2012imagenet}, VGG~\cite{simonyan2014very}, ResNet~\cite{he2016deep}, DenseNet~\cite{huang2017densely}, MobilNet~\cite{howard2017mobilenets}, ShuffleNet~\cite{zhang2018shufflenet}, ResNeXt~\cite{xie2017aggregated}, HRNet~\cite{wang2020deep}, RegNet~\cite{radosavovic2020designing}, ViT~\cite{dosovitskiy2020image}, SwinTransformer~\cite{liu2021swin}, CMT~\cite{guo2022cmt}, ConFormer~\cite{peng2021conformer}, CvT~\cite{wu2021cvt}) inherently encode rich semantic feature representations, directly facilitating the MedISeg capacity. At the same time, some certain elaborate feature regulation methods (\eg, lateral connection~\cite{badrinarayanan2017segnet}, residual mapping~\cite{he2016deep,huang2017densely}, encoder-decoder scheme~\cite{ronneberger2015u,guan20223d}, dense connection~\cite{quan2022centralized,li2018h}, feature pyramid~\cite{lin2017feature,wang2023coupling}, and the global context aggregation~\cite{wang2018non,zhang2022unabridged}) can boost the recognition performance. The integration of these sophisticated elements is the main reason that the MedISeg system performs so well. 
Besides, some training strategies (\eg, co-training~\cite{nigam2000analyzing,peng2020deep}, co-teaching~\cite{han2018co,robinet2022weakly}, co-learning~\cite{song2018collaborative,wei2020temporal}, and test-time-training~\cite{bartler2022mt3,sun2020test}) and loss functions (\eg, dice loss and Lovasz-softmax loss~\cite{yeung2022unified,salehi2017tversky}) are also indispensable components that can affect the MedISeg performance~\cite{athanasiadis2007semantic,ma2021loss}.
 
However, signs of progress are not proposed solely, they are usually mixed with the existing methods~\cite{steiner2021train,he2019bag}, leading to an extremely complex MedISeg system. In particular, at the present moment, an unabridged MedISeg system is usually composed of a large number of implementation details (including some non-learning model-agnostic pre-processing procedures) to achieve the desirable state-of-the-art performance~\cite{shen2017deep,shi2023transformer,zhang2022graph,bilic2019liver,zhang2019cascaded,shi2023discrepancy,li2021dual}. 
Unfortunately, in the current complex MedISeg systems, some marginal implementation strategies (also known as ``tricks'') are often disregarded, despite their significant impact on the system's performance. For example, as illustrated in Figure~\ref{fig1}, in the modified input stem of the prevalent ResNet~\cite{he2016deep} architecture (which is commonly treated as a prevailing backbone network for an MedISeg model), three cumulated $3 \times 3$ convolutional layers (in Figure~\ref{fig1} (b)) are used to replace the original $7 \times 7$ convolutional layer (in Figure~\ref{fig1} (a)) in the input stem to reduce the computational costs~\cite{zhao2017pyramid,chen2018encoder,hu2018squeeze}. Although this subtle change can significantly improve accuracy~\cite{he2019bag,touvron2021training,zhang2023augmented,ding2021repvgg,huang2017densely}, few publications explicitly mention this. Therefore, result comparisons of the performance based on such a modified implementation and the performance based on the original implementation are inherently unfair. 

 \begin{figure*}[t]
\centering
\includegraphics[width=.96\textwidth]{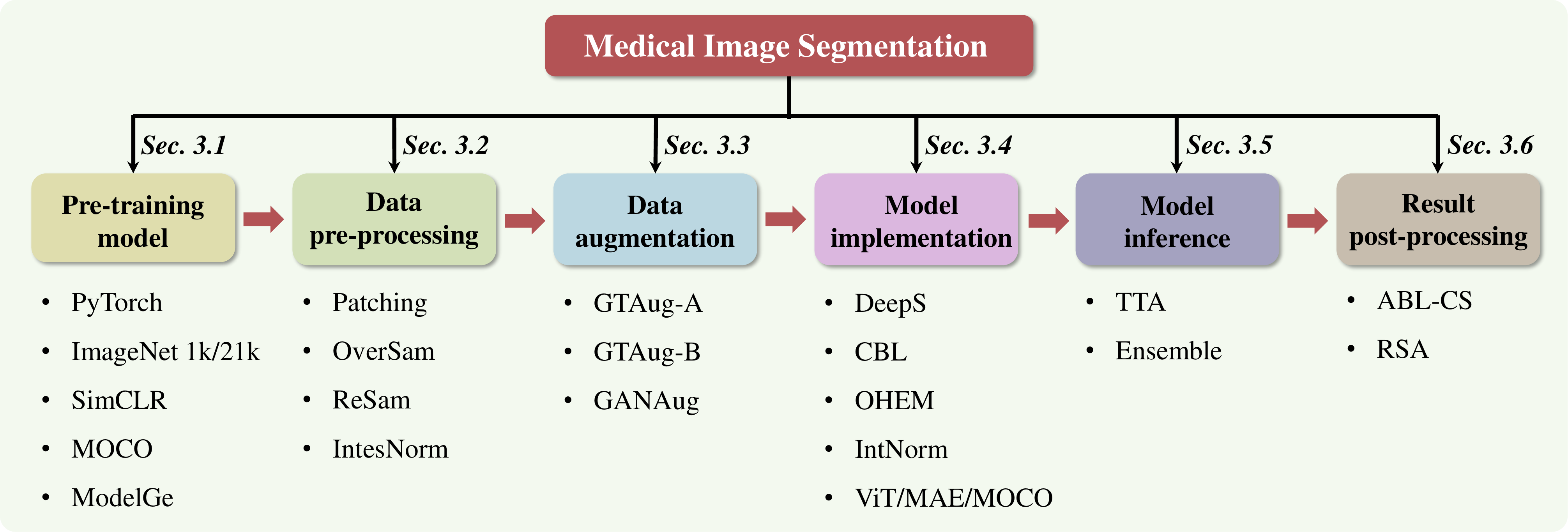}
\vspace{-4mm}
\caption{An illustration of the surveyed tricks and their latent relations. We separate a MedISeg system into six implementation phases, which are pre-training model, data pre-processing, data augmentation, model implementation, model inference, and result post-processing. For each phase of tricks, we experimentally explore their effectiveness on two typical semantic segmentation baselines, namely the 2D-UNet~\cite{ronneberger2015u} and 3D-UNet~\cite{cciccek20163d} with the help of four medical image segmentation datasets, \ie, 2D ISIC 2018~\cite{codella2019skin}, 2D CoNIC~\cite{graham2021conic}, 3D KiTS19~\cite{heller2021state}, and 3D LiTS~\cite{bilic2019liver}.}
\vspace{-4mm}
\label{fig2}
\end{figure*} 
The devil is in the details. In this work, to reveal effects of tricks on an MedISeg model, as illustrated in Figure~\ref{fig2}, according to a complete set of implementation phases including pre-training model (\textit{ref.}~Sec.~\ref{sec:3.1}), data pre-processing (\textit{ref.}~Sec.~\ref{sec:3.2}), data augmentation (\textit{ref.}~Sec.~\ref{sec:3.3}), model implementation (\textit{ref.}~Sec.~\ref{sec:3.4}), model inference (\textit{ref.}~Sec.~\ref{sec:3.5}), and result post-processing (\textit{ref.}~Sec.~\ref{sec:3.6}), we first collect a series of representative tricks that are easily overlooked in the current MedISeg models. 
Then, the effectiveness of these tricks is experimentally explored on consistent segmentation baselines including the typical 2D-UNet~\cite{ronneberger2015u} and 3D-UNet~\cite{cciccek20163d} with the help of the representative backbones, such that the influence of model variants (\ie, performance changes due to model changes) can be avoided. Compared to existing paper-driven technical surveys that only blandly focus on the advantage and limitation analyses of the image segmentation model, our work provides a large number of solid experimental results and is more technically operable for future work. Based on extensive experimental results on four medical image datasets (\ie, the challenging 2D ISIC 2018 lesion boundary segmentation dataset~\cite{codella2019skin}, 2D colon nuclei identification and counting challenge dataset~\cite{graham2021lizard,graham2021conic}, 3D kidney tumor segmentation 2019 dataset~\cite{heller2021state}, and 3D liver tumor segmentation challenge dataset~\cite{bilic2019liver}), we explicitly clarify the effect of these tricks. 
Moreover, based on the surveyed tricks and the used baseline models, we also open-sourced a strong MedISeg repository, where each of its component has the advantage of plug-and-play. 
It is believed that this milestone work not only completes a comprehensive technological survey of the state-of-the-art MedISeg approaches, but also offers a practical guide for addressing future medical image processing (especially the dense image predicted tasks) challenges including small dataset learning, class imbalance learning, multi-modality learning, and domain adaptation.

The main contributions are summarized as follows:
\begin{itemize}
\item We collect a series of fundamental MedISeg tricks for different implementation phases, and experimentally explore the effectiveness of these tricks on consistent baseline models.
\item We explicitly clarify the effectiveness of these tricks, and a large number of solid experimental results on both 2D and 3D medical image datasets compensate for the implementation neglect in MedISeg. 
\item We open-sourced a strong MedISeg repository, which includes rich segmentation tricks and each trick has the advantage of plug-and-play.
\item This milestone work will facilitate subsequent efforts to compare the experimental results of the MedISeg model under a fair environment.
\item This work will provide practical guidance for a wide range of medical image processing especially semantic segmentation challenges in the future.
\end{itemize}
\section{Preliminaries}
\label{sec:2}
\subsection{Baseline Models}
\label{sec:2.1}
In this work, ensure the comprehensiveness of the experiment, we choose the commonly used and representative 2D-UNet~\cite{ronneberger2015u} and 3D-UNet~\cite{cciccek20163d} as our baseline models. Details of these two baselines are as follows:

\subsubsection{2D-UNet} The 2D-UNet is a widely used architecture for medical image segmentation~\cite{cao2021swin,zhou2019review,sinha2020multi,zhang2021transfuse,feng2020cpfnet}. It consists of an encoder network and a decoder network. The encoder network follows the fully convolutional architecture and comprises four spatial-tapering stages, each consisting of two $3 \times 3$ convolutional layers followed by a rectified linear unit (ReLU) activation function~\cite{nair2010rectified,agarap2018deep} and a global max pooling layer (with stride size $S=2$). The decoder network, which takes the output of the encoder network as input, also has four stages that correspond to the same spatial encoder stages. Each decoder stage uses a 2D transposed convolutional operator to upsample feature maps $2\times$ via bilinear interpolation operation~\cite{smith1981bilinear}. Then, two $3 \times 3$ convolutional layers and a ReLU activation function\cite{nair2010rectified,agarap2018deep} are deployed in sequence. In the final decoder network layer, the channel size of the output feature maps is assigned to the class size of the used dataset via a $2 \times 2$ convolutional layer.

\subsubsection{3D-UNet} The 3D-UNet shares a similar network architecture with the 2D-UNet~\cite{ronneberger2015u}, consisting of a 3D encoder network and a 3D decoder network~\cite{hatamizadeh2022unetr,xie2021cotr,yu2020c2fnas,wang2019volumetric,zhang2020inter,liao2020iteratively}. It is commonly used for 3D image segmentation~\cite{wu2022d,yan2022after,hatamizadeh2022unetformer}. The primary differences between 3D-UNet and 2D-UNet are: \romannumeral1) the 2D convolution layer is replaced with a 3D convolution layer~\cite{ji20123d}; \romannumeral2) a lateral connection is added between the same level of the encoder and decoder stages, with the same spatial and channel size; and \romannumeral3) input image intensity normalization is applied using a normalization layer~\cite{ioffe2015batch}~\cite{nair2010rectified,agarap2018deep}.

\subsection{Experimental Datasets}
\label{sec:2.2}
To ensure a comprehensive experimental evaluation and avoid bias towards characteristics of a specific dataset, we selected four representative medical image datasets in this paper. These include a 2D dataset with a common object size, namely the ISIC 2018 Lesion Boundary Segmentation dataset~\cite{codella2019skin}; a 2D dataset with small object size, the Colon Nuclei Identification and Counting Challenge (CoNIC) dataset~\cite{graham2021lizard,graham2021conic,graham2019hover}; the 3D Kidney Tumor Segmentation 2019 (KiTS19) dataset~\cite{heller2021state}; and the 3D Liver Tumor Segmentation Challenge (LiTS) dataset~\cite{bilic2019liver}.
Details of each dataset are introduced as follows:

\subsubsection{2D ISIC 2018} The ISIC 2018 dataset~\cite{codella2019skin} is a challenging yet representative 2D skin lesion boundary segmentation dataset in the computer-aided diagnosis domain. It consists of $2,594$ JPEG dermoscopic images and $2,594$ PNG ground truth (GT) images, with one or more lesion regions of varying sizes in each image. Each skin lesion image has a uniform spatial size of $600 \times 450$. We only need to segment two categories of regions in this dataset, namely the foreground ``lesion'' region and the ``background''. For our experiments, we randomly split the dataset into $80\%$ training and $20\%$ test sets, as done in previous studies~\cite{abraham2019novel,azad2020attention}. In each cross-validation, we further randomly split $10\%$ of the training set as the validation set.

\subsubsection{2D CoNIC} The 2D CoNIC challenge dataset comprises images from the Lizard dataset~\cite{graham2021lizard}, which includes six nuclear categories (\ie, ``epithelial cells'', ``connective tissue cells'', ``lymphocytes'', ``plasma cells'', ``neutrophils'', and ``eosinophils''), and one ``background''. Each original image in the Lizard dataset~\cite{graham2021lizard,graham2021conic,graham2019hover} has a ground truth label that contains information about the instance map, nuclear categories, bounding boxes segmentation mask, and nucleus counts. For our experiments, we use the segmentation mask for image segmentation and only distinguish between the foreground object ``nuclei'' and the ``background'', resulting in a binary segmentation task as in~\cite{codella2019skin}. The dataset provides $4,981$ image patches in RGB format with a size of $256\times256$. Following previous studies~\cite{weigert2022nuclei,bohland2022ciscnet,lee2022using}, we randomly split all images into $80\%$ for training and $20\%$ for testing. In each cross-validation, we further randomly split $10\%$ of the training set as the validation set. 

\subsubsection{3D KiTS19} The KiTS19 dataset provides $210$ high-quality annotated 3D abdominal computed tomography images of patients, consisting of three categories: ``kidney'', ``tumor'', and ``background''. The positions of the kidney and tumor are relatively fixed in the given images. For our experiments, we follow the commonly used binary segmentation settings~\cite{heller2021state,heller2019kits19,isensee2021nnu}: settings-\romannumeral1) the foreground ``kidney'' and the ``background''; settings-\romannumeral2) the foreground "tumor" and the ``background''. Although KiTS19 provides other human structures (\eg, ureters, arteries, and veins), they are not in the range that needs to be segmented and are uniformly considered as the background. Each image and its corresponding ground truth mask in KiTS19 are provided in the NIFTI format, including the number of slices (averaging $216$ slices), height ($512$), and width ($512$). To ensure ground truth quality, we follow previous studies~\cite{isensee2019attempt,heller2021state} and remove cases $15$, $23$, $37$, $68$, $125$, and $133$ from the original dataset, leaving $204$ cases for training and testing. In each cross-validation, we randomly split off $10\%$ from the training set as the validation set, as done in~\cite{heller2019kits19,heller2021state}. Although the dataset's official website provides an additional $90$ test cases, the ground truth masks are not publicly accessible, so we exclude these cases from our experiments.

\subsubsection{3D LiTS} Liver cancer is a common tumor, primarily affecting men~\cite{bilic2019liver,li2020attention,li2020transformation}. To advance the automatic segmentation of lesions, LiTS proposes a benchmark based on contrast-enhanced abdominal computed tomography scans. LiTS contains 130 scans for training and 70 scans for testing, with images initially provided in the PNG format and a spatial size of $256 \times 256$. Each image contains two categories: ``background'' and ``liver''. We treat the tumor area and the liver area equivalently, following~\cite{gul2022deep,kushnure2022hfru}. During training, all images are clipped to a spatial size of $200 \times 200$. In each cross-validation, we randomly split off $10\%$ from the training set as the validation set, as done in~\cite{li2020transformation}.

\subsection{Experimental Settings}
\label{sec:2.3}
\myparagraph{Platform.} All experiments in our work are conducted using the PyTorch deep learning platform~\cite{paszke2019pytorch} with NVIDIA GeForce RTX 2080 GPUs. To minimize potential coding errors, we utilize PyTorch's official functions to perform the necessary operations, including basic operations, loss calculations, and metric calculations.

\myparagraph{Backbone.} Due to the dataset size and network maturity, we employ the widely recognized ResNet-50~\cite{he2016deep} as the default backbone network, which follows the classic implementation of the fully convolutional network architecture~\cite{long2015fully,zhang2020causal,li2018fully}. We note that the backbone network's input stem is implemented using the original cumulative convolution method, as depicted in Figure~\ref{fig1} (a).

\myparagraph{Evaluation metrics.} We employ the commonly used Recall, Precision, Dice~\cite{milletari2016v}, and Intersection over Union (IoU) as our primary metrics~\cite{heller2021state,isensee2019attempt,isensee2021nnu,estienne2020deep}. In particular, we perform five-fold cross-validation to evaluate the model performance and report the average result.

\subsubsection{Implementation Details of 2D-UNet} Following the previous work in~\cite{abraham2019novel,azad2020attention}, 2D training images in ISIC 2018~\cite{codella2019skin} and in CoNIC~\cite{graham2021lizard,graham2021conic,graham2019hover} are uniformly resized into the same fixed input size of $200 \times 200$, and are also normalized using the mean value and the standard deviation of the ImageNet dataset~\cite{deng2009imagenet} as in~\cite{zhang2021transfuse,salehi2017tversky,shin2018medical}. The initial learning rate is set to $0.0003$. The adaptive moment estimation~\cite{kingma2014adam} is used as the optimizer, where the weight decay is set to $0.0005$. The whole model is trained with $200$ training epochs with a batch size of $32$, and the results of the best model on the validation set are used for testing as in~\cite{yeung2022unified,isensee2021nnu}. Following~\cite{li2018h,bilic2019liver}, we employ the commonly used pixel-level cross-entropy~\cite{drozdzal2016importance} loss as the default loss function. 

\subsubsection{Implementation Details of 3D-UNet} Following the commonly used experimental setting~\cite{heller2021state,isensee2019attempt,santini2019kidney,estienne2020deep,bilic2019liver,li2020attention,li2020transformation} for KiTS19~\cite{heller2021state} and LiTS~\cite{bilic2019liver}, Hounsfield units are clipped to $[-79,..., 304]$ and the voxel spacing is resampled by the coefficient of $3.22 \times 1.62 \times 1.62$ mm$^3$ as in~\cite{heller2019kits19,isensee2021nnu,yeung2022unified}. 
We down-sample the computed tomography scans on the cross-section and resample it to adjust the z-axis spacing of all data to 1mm for LiTS~\cite{bilic2019liver}.
The training voxel size (patch size) is set to $96 \times 96 \times 96$, which is just right to be crammed into an NVIDIA GeForce RTX 2080 GPU. The stochastic gradient descent~\cite{ketkar2017stochastic} is used as the optimizer, where the momentum is set to $0.9$ and the weight decay is set to $0.0001$, respectively. The initial learning rate is set to $0.01$, and the batch size is set to $2$. The model is trained in an end-to-end fashion with $100$ epochs, and the results of the best model on the validation set are used for testing as in~\cite{yeung2022unified,isensee2021nnu}. The cross-entropy loss is used as the unique loss function as in~\cite{qayyum2020automatic,santini2019kidney}. In the inference phase, we perform patch-wise overlap with 50\% region overlap (\ie, $48 \times 48 \times 48$).
\section{Methods and Experiments}
\label{sec:3}
We divide the complete MedISeg system into six phases, with each phase exploring a set of representative tricks empirically. We note that all experiments are conducted using the preliminary settings described in Section~\ref{sec:2.3}, without any specific modifications mentioned.
\subsection{Pre-Training Model}
\label{sec:3.1}
Pre-training models can provide favorable parameters~\cite{wang2021dense,liu2021swin}, but the impact of different pre-training models is usually overlooked. For instance, when using the popular ImageNet~\cite{deng2009imagenet} pre-training model, authors frequently state that their network is pre-trained on the ImageNet, without specifying the particular pre-training implementation. However, pre-training on ImageNet can take at least two basic forms (\ie, 1k and 21k versions), and fair result comparisons require detailed pre-training implementation explanations. To this end, in this subsection, we explore six commonly used and publicly available pre-trained weights on MedISeg. These pre-trained weights can be divided into the following two main categories: fully-supervised (\ie, PyTorch official weights~\cite{paszke2019pytorch} and model-oriented ImageNet-1k/21k weights~\cite{deng2009imagenet}) and self-supervised (\ie, SimCLR weights~\cite{chen2020simple}, MoCo weights~\cite{he2020momentum}, and Model Genesis (ModelGe) weights~\cite{zhou2019models}).

\subsubsection{PyTorch Official Weights}
In the PyTorch repository, there are some backbone pre-trained weights provided by \verb"torchvision.models"~\cite{paszke2019pytorch}. These weights are obtained by training the corresponding backbone network for the single-label image classification task on ImageNet-1k dataset~\cite{deng2009imagenet}, where ``1k'' denotes that this dataset consists of $1,000$ classes of the common scenes.

\subsubsection{Model-Oriented ImageNet-1k Weights} 
Besides PyTorch official weights, the model creator usually releases pre-trained weights on ImageNet~\cite{deng2009imagenet,yalniz2019billion} as well. For example, the model-oriented ImageNet-1k weights can be obtained by training ResNet-50~\cite{he2016deep} for image classification on the ImageNet-1k dataset. We then use the obtained trained weights for downstream vision tasks.

\subsubsection{Model-Oriented ImageNet-21k Weights} 
Compared to ImageNet-1k~\cite{deng2009imagenet}, ImageNet-21k is a more general and comprehensive dataset version, which has about $21,000$ object classes in total~\cite{hendrycks2019using} for image classification. Therefore, the trained weights on ImageNet-21k are also conducive to downstream tasks~\cite{dosovitskiy2020image,liu2021swin,kolesnikov2020big}.
\begin{table*}[t]
\begin{center}
\renewcommand\arraystretch{1.5}
\setlength{\tabcolsep}{5pt}{
\caption{Experimental results on different pre-trained weights. ``PyTorch'',  ``Imag-1k'', ``Imag-21k'', ``SimCLR'', ``MOCO'' and ``ModelGe'' denotes the PyTorch~\cite{paszke2019pytorch} official weights, model-oriented ImageNet 1k~\cite{hendrycks2019using} weights, model-oriented ImageNet 21k~\cite{hendrycks2019using} weights, SimCLR~\cite{chen2020simple} weights, MOCO~\cite{he2020momentum} weights and the model genesis~\cite{zhou2019models} weights, respectively. ``+'' means fine-tuning the baseline model on the corresponding weights.}
\vspace{-3mm}
\begin{tabular}{ r | c c c c | c c c c} 
 Methods & Recall (\%) & Percision (\%) & Dice (\%) & IoU (\%) & Recall (\%) & Percision (\%) & Dice (\%) & IoU (\%) \\ 
\hline \hline 
\multirow{2}{*}{2D-UNet~\cite{ronneberger2015u}} & \multicolumn{4}{c}{ISIC 2018~\cite{codella2019skin}} & \multicolumn{4}{c}{CoNIC~\cite{graham2021lizard}} \\
\cline{2-9}
~ & 88.18 & 89.88 & 86.89 & 85.80 & 78.12 & 77.25 & 77.23 & 77.58 \\ 
\cdashline{1-9}[0.8pt/2pt]
+ PyTorch~\cite{paszke2019pytorch} & 89.28$_{\color{red}{+1.10}}$ & 90.08$_{\color{red}{+0.20}}$ & 88.09$_{\color{red}{+1.20}}$  & 87.07$_{\color{red}{+1.27}}$ & 78.08$_{\color{blue}{-0.04}}$ & 79.21$_{\color{red}{+1.96}}$ & 78.08$_{\color{red}{+0.85}}$  & 78.38$_{\color{red}{+0.80}}$ \\
+ Imag-1k~\cite{hendrycks2019using} & 89.19$_{\color{red}{+1.01}}$ & 90.07$_{\color{red}{+0.19}}$ & 87.99$_{\color{red}{+1.10}}$ & 86.93$_{\color{red}{+1.13}}$ & 79.48$_{\color{red}{+1.36}}$ & 78.69$_{\color{red}{+1.44}}$ & 78.70$_{\color{red}{+1.47}}$ & 78.77$_{\color{red}{+1.19}}$ \\
+ Imag-21k~\cite{hendrycks2019using} & 90.21$_{\color{red}{+2.03}}$ & 91.48$_{\color{red}{+1.60}}$ & 89.38$_{\color{red}{+2.49}}$ & 88.00$_{\color{red}{+2.20}}$ & 78.79$_{\color{red}{+0.67}}$ & 79.66$_{\color{red}{+2.41}}$ & 78.75$_{\color{red}{+1.52}}$ & 78.91$_{\color{red}{+1.33}}$ \\
+ SimCLR~\cite{chen2020simple} & 88.09$_{\color{blue}{-0.09}}$ & 89.93$_{\color{red}{+0.05}}$ & 86.95$_{\color{red}{+0.06}}$  & 85.90$_{\color{red}{+0.10}}$ & 77.87$_{\color{blue}{-0.25}}$ & 77.44$_{\color{red}{+0.19}}$ & 77.17$_{\color{blue}{-0.06}}$ & 77.53$_{\color{blue}{-0.05}}$ \\ 
+ MOCO~\cite{he2020momentum} & 87.99$_{\color{blue}{-0.19}}$ & 90.11$_{\color{red}{+0.23}}$ & 86.88$_{\color{blue}{-0.01}}$ & 85.84$_{\color{red}{+0.04}}$ & 77.98$_{\color{blue}{-0.14}}$ & 77.65$_{\color{red}{+0.40}}$ & 77.35$_{\color{red}{+0.12}}$  & 77.67$_{\color{red}{+0.09}}$ \\ 
\hline \hline 
\multirow{2}{*}{3D-UNet~\cite{cciccek20163d}} & \multicolumn{4}{c}{KiTS19~\cite{heller2021state}:~settings-\romannumeral1} & \multicolumn{4}{c}{KiTS19~\cite{heller2021state}:~settings-\romannumeral2} \\
\cline{2-9}
~ & 91.01 & 95.20 & 92.50 & 87.35 & 27.35 & 46.71 & 29.63 & 21.51 \\ 
\cdashline{1-9}[0.8pt/2pt]
+ ModelGe~\cite{zhou2019models} & 90.70$_{\color{blue}{-0.31}}$ & 95.48$_{\color{red}{+0.28}}$ & 92.29$_{\color{blue}{-0.21}}$ & 87.18$_{\color{blue}{-0.17}}$ & 28.17$_{\color{red}{+0.82}}$ & 47.62$_{\color{red}{+0.91}}$ & 29.95$_{\color{red}{+0.32}}$ & 21.75$_{\color{red}{+0.24}}$
\end{tabular}
\label{tab1}}
\vspace{-4mm}
\end{center}
\end{table*}
\begin{table}[t]
\begin{center}
\renewcommand\arraystretch{1.5}
\setlength{\tabcolsep}{2.5pt}{
\caption{Experimental results on 3D LiTS dataset~\cite{bilic2019liver}. ``ModelGe'' denotes the model genesis~\cite{zhou2019models} weight. ``+'' means fine-tuning the baseline model on the corresponding pre-trained weights.}
\vspace{-3mm}
\begin{tabular}{ r | c c c c} 
 Methods & Recall (\%) & Percision (\%) & Dice (\%) & IoU (\%) \\ 
\hline \hline 
3D-UNet~\cite{cciccek20163d} & 89.33 & 84.03 & 86.11 & 76.44 \\ 
\cdashline{1-5}[0.8pt/2pt]
+ ModelGe~\cite{zhou2019models} & 90.54$_{\color{red}{+1.21}}$ & 84.66$_{\color{red}{+0.63}}$ & 86.99$_{\color{red}{+0.88}}$ & 77.67$_{\color{red}{+1.23}}$  
\end{tabular}
\label{tab2}}
\vspace{-4mm}
\end{center}
\end{table}

\subsubsection{SimCLR Weights} 
SimCLR~\cite{chen2020simple} demonstrates that using a learnable nonlinear transformation between feature representations and the contrastive learning loss can improve feature representations~\cite{grill2020bootstrap,jing2020self}. To this end, SimCLR mainly consists of three steps: 1) the input image is grouped into some image patches; 2) different data augmentation strategies are implemented on image patches for different batches; 3) the model is trained to obtain the similar results for the same image patches with different augmentations, and mutually exclude other results. In our work, we use the SimCLR weights that are trained on ImageNet-1k~\cite{deng2009imagenet} for classification~\cite{he2016deep}.

\subsubsection{MoCo Weights} 
MoCo~\cite{he2020momentum} is one of the classical self-supervised contrastive learning methods. It aims to address the problem of sampled feature inconsistency in the memory bank~\cite{chen2020generative,chen2021exploring}. To this end, MoCo uses a queue to store and sample the negative samples; that is, it stores feature vectors of multiple recent batches used for training. In its implementation, the fixed network remains unchanged and a linear layer with the softmax layer is added to the end of the backbone for classification in an unsupervised training manner. In our work, we use the MoCo weights that are trained on ImageNet-1k~\cite{deng2009imagenet} for classification on ResNet-50~\cite{he2016deep}.

\subsubsection{ModelGe Weights} 
ModelGe~\cite{zhou2019models} is an advanced self-supervised model pre-training technology, which usually consists of four transformation operations (\ie, non-linear, local pixel shuffling, out-painting, and in-painting) for single image restoration on computed tomography and magnetic resonance imaging images~\cite{rajwade2012image,taleb2022contig}. 
In its implementation, the network is trained to learn a general visual representation by restoring the original image patch from the transformed one, where the weight for transformation operations is set to $0.9$ for non-linear, $0.5$ for local pixel shuffling, $0.8$ for out-painting, and $0.2$ for in-painting following~\cite{zhao2021anomaly,haghighi2021transferable,li2021rotation}. Based on the trained model, the obtained trained weights can be used as the pre-trained weights for downstream models.

\subsubsection{Experimental Results}
Table~\ref{tab1} shows our experimental results on ISIC 2018~\cite{codella2019skin}, CoNIC~\cite{graham2021conic}, and KiTS19~\cite{kutikov2009renal}, while Table~\ref{tab2} presents our results on LiTS~\cite{bilic2019liver}. \textbf{Overall, we observe that the fine-tuned model outperforms the baseline model trained from scratch.} This observation validates the effectiveness of pre-trained weights and confirms that different pre-trained weights have varying effects. For instance, compared to the baseline results of 2D-UNet~\cite{ronneberger2015u} on ISIC 2018, pre-trained weights enhance performance across almost all evaluation metrics. Specifically, fine-tuning using ImageNet-21k~\cite{deng2009imagenet} pre-trained weights results in a maximum performance gain of $2.03\%$ Recall, $1.60\%$ Precision, $2.49\%$ Dice, and $2.20\%$ IoU, respectively. These gains demonstrate the powerful representation ability of ImageNet-21k~\cite{zheng2021rethinking,liu2021swin}. 
Moreover, the performance gain of the pre-trained model on Precision is relatively small, with an average gain of only 0.45\%. We observe similar experimental conclusions on the CoNIC~\cite{graham2021conic} dataset. Specifically, the performance gains on Recall are relatively small (with an average gain of only 0.41\%), whereas the performance gains on Precision are relatively large (with an average gain of 1.08\%). 
Therefore, \textbf{in 2D medical image segmentation, pre-training using ImageNet-21k is the best choice.} Although the model's performance drops on some evaluation metrics, we believe that this is not due to the pre-trained weights. The pre-trained weights are obtained from natural scenes, while our task is about medical images, resulting in a domain gap. This problem can be addressed in the future by using pre-trained weights on medical images.

Comparing the fine-tuned 3D-UNet using the released ModelGe weights to the baseline results on KiTS19~\cite{kutikov2009renal} and LiTS~\cite{bilic2019liver}, we observe that using the ModelGe weights can result in a maximum performance gain of $1.21\%$, $0.91\%$, $0.88\%$, and $1.23\%$ in Recall, Precision, Dice, and IoU, respectively, on setting-\romannumeral2~(\ie, segment the foreground ``tumor'' and the "background"). However, on setting-\romannumeral1~on KiTS19~\cite{kutikov2009renal}, we observe that three-quarters of the performance has a slight drop, with $-0.31\%$ Recall, $-0.21\%$ Dice, and $-0.17\%$ IoU. We speculate that this may be because the kidney region in KiTS19~\cite{kutikov2009renal} is more sensitive to the initialized parameters. These 3D experimental results demonstrate that \textbf{if there are no better pre-training weights to choose from, ModelGe weights can still achieve satisfactory results overall.} Moreover, this observation inspires us to consider not only the differences between the network architectures of the fine-tuned models but also the state of the used dataset as critical factors~\cite{he2019bag,bao2020unilmv2}.
\subsection{Data Pre-Processing}
\label{sec:3.2}
Due to the specific data characteristics of 3D medical images (\eg, modalities and resolutions), pre-processing is necessary to achieve satisfactory segmentation performance~\cite{ahmed2002modified,zhang2001segmentation}. In this work, we investigate the effectiveness of four commonly used image pre-processing techniques in 3D-UNet~\cite{cciccek20163d}: patching~\cite{lin2022label}, oversampling (OverSam)~\cite{kotsiantis2006handling}, resampling (ReSam)~\cite{pizer2003deformable}, and intensity normalization (IntesNorm)~\cite{isensee2021nnu}.

\subsubsection{Patching} 
Some medical images, such as MRI~\cite{imai2019high} and pathology images~\cite{lin2022label}, can be very large in spatial size and lack sufficient training samples, making it impractical to train a MedISeg model directly on these images~\cite{hesamian2019deep,mainak2019state}. Instead, it is common to resample the entire image into smaller image patches at different spatial scales, with or without overlaps. This approach reduces GPU memory requirements and enables more effective model training. The patch size is one of the most critical factors affecting model performance. In our work, following the experimental settings on 3D-UNet~\cite{yeung2022unified,isensee2021nnu}, we set the training patch size to $96 \times 96 \times 96$ without overlap and the patch size to $96 \times 96 \times 96$ with a $50\%$ region overlap during the inference phase.
\begin{table*}[t]
\begin{center}
\renewcommand\arraystretch{1.5}
\setlength{\tabcolsep}{5pt}{
\caption{Experimental results on data pre-processing tricks, where the two columns of results are experiments of setting-$\textrm{\romannumeral1}$ and setting-$\textrm{\romannumeral2}$ (as introduced in section~\ref{sec:2.2}), respectively. ``OverSam'', ``ReSam'', ``IntesNorm'' denotes Oversampling~\cite{kotsiantis2006handling}, Resampling~\cite{pizer2003deformable}, and Intensity normalization~\cite{isensee2021nnu}, respectively. ``/o'' denotes that this trick is not implemented under this setting, and ``\emph{NaN}'' denotes that the model under this setting is corrupted.}
\vspace{-3mm}
\begin{tabular}{r | c c c c | c c c c} 
Methods & Recall (\%) & Percision (\%) & Dice (\%) & IoU (\%) & Recall (\%) &  Percision (\%) & Dice (\%) & IoU (\%) \\ 
\hline \hline 
\multirow{2}{*}{3D-UNet~\cite{cciccek20163d}} & \multicolumn{4}{c}{KiTS19~\cite{heller2021state}:~settings-\romannumeral1} & \multicolumn{4}{c}{KiTS19~\cite{heller2021state}:~settings-\romannumeral2} \\
\cline{2-9}
~ & 91.01 & 95.20 & 92.50 & 87.35 & 27.35 & 46.71 & 29.63 & 21.51  \\
\cdashline{1-9}[0.8pt/2pt]
Patching$_{32}$~\cite{isensee2021nnu} & \emph{NaN} & \emph{NaN} & \emph{NaN} & \emph{NaN} & \emph{NaN} & \emph{NaN} & \emph{NaN} & \emph{NaN} \\
Patching$_{64}$~\cite{isensee2021nnu} & 37.38$_{\color{blue}{-53.63}}$ & 59.51$_{\color{blue}{-35.69}}$ & 43.47$_{\color{blue}{-49.03}}$ & 29.87$_{\color{blue}{-57.48}}$ & 5.91$_{\color{blue}{-21.44}}$ & 9.40$_{\color{blue}{-37.31}}$ & 5.07$_{\color{blue}{-24.56}}$ & 2.89$_{\color{blue}{-18.62}}$  \\
Patching$_{96}$~\cite{isensee2021nnu} & 91.01$_{\color{red}{+0.00}}$ & 95.20$_{\color{red}{+0.00}}$ & 92.50$_{\color{red}{+0.00}}$ & 87.35$_{\color{red}{+0.00}}$ & 27.35$_{\color{red}{+0.00}}$ & 46.71$_{\color{red}{+0.00}}$ & 29.63$_{\color{red}{+0.00}}$ & 21.51$_{\color{red}{+0.00}}$ \\
Patching$_{128}$~\cite{isensee2021nnu} & 89.37$_{\color{blue}{-1.64}}$ & 95.24$_{\color{red}{+0.04}}$ & 91.56$_{\color{blue}{-0.94}}$ & 85.88$_{\color{blue}{-1.47}}$ & 37.81$_{\color{red}{+10.46}}$ & 52.28$_{\color{red}{+5.57}}$ & 38.06$_{\color{red}{+8.43}}$ & 28.81$_{\color{red}{+7.30}}$ \\
Patching$_{160}$~\cite{isensee2021nnu} & 92.85$_{\color{red}{+1.84}}$ & 95.47$_{\color{red}{0.27}}$ & 93.79$_{\color{red}{+1.29}}$ & 89.32$_{\color{red}{+1.97}}$ & 38.42$_{\color{red}{+11.07}}$ & 57.39$_{\color{red}{+10.68}}$ & 39.29$_{\color{red}{+9.66}}$ & 29.71$_{\color{red}{+8.20}}$ \\
Patching$_{192}$~\cite{isensee2021nnu} & 92.77$_{\color{red}{+1.76}}$ & 95.02$_{\color{blue}{-0.18}}$ & 93.44$_{\color{red}{+0.94}}$ & 88.82$_{\color{red}{+1.47}}$ & 40.32$_{\color{red}{+12.97}}$ & 57.22$_{\color{red}{+10.51}}$ & 40.36$_{\color{red}{+10.73}}$ & 30.83$_{\color{red}{+9.32}}$ \\
\cdashline{1-9}[0.8pt/2pt]
+ OverSam~\cite{kotsiantis2006handling} & 91.85$_{\color{red}{+0.84}}$ & 95.05$_{\color{blue}{-0.15}}$ & 92.93$_{\color{red}{+0.43}}$ & 88.08$_{\color{red}{+0.73}}$ & 35.46$_{\color{red}{+8.11}}$ & 53.69$_{\color{red}{+6.98}}$ & 35.69$_{\color{red}{+6.06}}$ & 26.56$_{\color{red}{+5.05}}$  \\
(/o) ReSam~\cite{pizer2003deformable} & 0.18$_{\color{blue}{-90.83}}$ & 25.49$_{\color{blue}{-69.71}}$ & 0.35$_{\color{blue}{-92.15}}$ & 0.18$_{\color{blue}{-87.17}}$ & \emph{NaN} & \emph{NaN} & \emph{NaN} & \emph{NaN} \\
(/o) IntesNorm~\cite{ioffe2015batch} & 90.23$_{\color{blue}{-0.78}}$ & 95.33$_{\color{red}{+0.13}}$ & 92.20$_{\color{blue}{-0.30}}$ & 86.65$_{\color{blue}{-0.70}}$ & 27.15$_{\color{blue}{-0.20}}$ & 49.29$_{\color{red}{+2.58}}$ & 29.63$_{\color{red}{+0.00}}$ & 21.73$_{\color{red}{+0.22}}$
\end{tabular}
\label{tab3}}
\vspace{-4mm}
\end{center}
\end{table*}


\subsubsection{OverSam} 
OverSam~\cite{kotsiantis2006handling} is proposed to address the issue of class imbalance between positive and negative samples in the minority class~\cite{krawczyk2019radial}. Currently, a group of OverSam schemes have been proposed, including random oversampling~\cite{kotsiantis2006handling}, synthetic minority oversampling (SMOTE)~\cite{chawla2002smote}, borderline SMOTE~\cite{han2005borderline}, and adaptive synthetic sampling~\cite{he2008adasyn}. Previous experimental results have shown that OverSam does not affect the model slope but can amplify the model intercept~\cite{fithian2014local,lemaitre2017imbalanced}. In our work, we adopt the prevailing OverSam scheme as proposed in~\cite{isensee2021nnu}, where $70\%$ of the selected training samples are from random image locations and $30\%$ of the image patches are guaranteed to contain at least one foreground class. This way, each training sample can simultaneously include one foreground image patch and one random image patch.

\subsubsection{ReSam} 
ReSam~\cite{pizer2003deformable} is proposed to improve the representational ability of the used dataset. Because the available sample ability is sometimes limited and heterogeneous, a better sub-sample dataset can be obtained via a random/nonrandom ReSam strategy~\cite{seppa2007high,lartizien2010comparison}. In its implementation, ReSam mainly consists of four steps: 1) spacing interpolation; 2) window transform; 3) acquisition of mask effective range, and 4) generation of sub-images. Based on the reorganized sub-sample dataset, a better-performing recognition model can be trained~\cite{pizer2003deformable,zhu2019boundary}. In our baseline implementation, the commonly used random ReSam strategy has been used. To demonstrate its importance in MedISeg, in this section, we explore the effect of the ReSam strategy by removing it (\ie, /o) in our experiments, \ie, the image pixels are directly interpolated and scaled, such that the actual distances represented by the pixels are the same.

\subsubsection{IntesNorm}
IntesNorm~\cite{shah2011evaluating} is a specific normalization strategy for medical images~\cite{collewet2004influence,shinohara2014statistical}. There are usually two commonly used IntesNorm methods: \textit{z}-scoring for all modalities and another one for computed tomography images~\cite{isensee2021nnu}. In our work, we mainly explore the effectiveness of IntesNorm by removing it (\ie, /o) on KiTS19~\cite{heller2021state} in our experiments. Following the common implementation~\cite{shah2011evaluating,isensee2021nnu}, a global normalization scheme is adopted in this paper, where $0.5\%$ of the foreground voxels is used for clipping and computing the foreground mean, and $99.5\%$ of the foreground voxels is used for computing the standard deviation. 
\begin{table}[t]
\begin{center}
\footnotesize
\renewcommand\arraystretch{1.5}
\setlength{\tabcolsep}{.3pt}{
\caption{Experimental results on 3D LiTS dataset~\cite{bilic2019liver} on data pre-processing tricks. ``OverSam'', ``ReSam'', ``IntesNorm'' denotes Oversampling~\cite{kotsiantis2006handling}, Resampling~\cite{pizer2003deformable}, and Intensity normalization~\cite{isensee2021nnu}, respectively. ``/o'' denotes that this trick is not implemented under this setting, and ``\emph{NaN}'' denotes that the model under this setting is corrupted.}
\vspace{-2mm}
\begin{tabular}{ r | c c c c} 
 Methods & Recall (\%) & Percision (\%) & Dice (\%) & IoU (\%) \\ 
\hline \hline 
3D-UNet~\cite{cciccek20163d} & 89.33 & 84.03 & 86.11 & 76.44  \\ 
\cdashline{1-5}[0.8pt/2pt]
Patching$_{32}$~\cite{isensee2021nnu} & \emph{NaN} & \emph{NaN} & \emph{NaN} & \emph{NaN} \\
Patching$_{64}$~\cite{isensee2021nnu} & 73.27$_{\color{blue}{-16.06}}$ & 78.57$_{\color{blue}{-5.46}}$ & 75.14$_{\color{blue}{-10.97}}$ & 60.71$_{\color{blue}{-15.73}}$ \\
Patching$_{96}$~\cite{isensee2021nnu} & 89.33$_{\color{red}{+0.00}}$ & 84.03$_{\color{red}{+0.00}}$ & 86.11$_{\color{red}{+0.00}}$ & 76.44$_{\color{red}{+0.00}}$ \\
Patching$_{128}$~\cite{isensee2021nnu} & 92.05$_{\color{red}{+2.72}}$ & 91.55$_{\color{red}{+7.52}}$ & 91.44$_{\color{red}{+5.33}}$ & 84.60$_{\color{red}{+8.16}}$  \\
Patching$_{160}$~\cite{isensee2021nnu} & 93.29$_{\color{red}{+3.96}}$ & 94.94$_{\color{red}{+10.91}}$ & 93.88$_{\color{red}{+7.77}}$ & 88.76$_{\color{red}{+12.32}}$ \\
Patching$_{192}$~\cite{isensee2021nnu} & 93.31$_{\color{red}{+3.98}}$ & 95.35$_{\color{red}{+11.32}}$ & 94.08$_{\color{red}{+7.97}}$ & 89.18$_{\color{red}{+12.74}}$ \\
\cdashline{1-5}[0.8pt/2pt]
+ OverSam~\cite{kotsiantis2006handling} & 89.78$_{\color{red}{+0.45}}$ & 87.55$_{\color{red}{+3.52}}$ & 88.14$_{\color{red}{+2.03}}$ & 79.85$_{\color{red}{+3.41}}$ \\
(/o) ReSam~\cite{pizer2003deformable} &  73.19$_{\color{blue}{-16.14}}$ & 78.86$_{\color{blue}{-5.17}}$ & 73.87$_{\color{blue}{-12.24}}$ & 60.84$_{\color{blue}{-15.60}}$  \\
(/o) IntesNorm~\cite{ioffe2015batch} & 87.64$_{\color{blue}{-1.69}}$ & 81.14$_{\color{blue}{-2.89}}$ & 83.68$_{\color{blue}{-2.43}}$ & 72.37$_{\color{blue}{-4.07}}$  \\
\end{tabular}
\label{tab4}}
\vspace{-4mm}
\end{center}
\end{table}

\subsubsection{Experimental Results}
Table~\ref{tab3} and Table~\ref{tab4} present our experimental results on KiTS19~\cite{kutikov2009renal} and LiTS~\cite{bilic2019liver}, respectively. We observe that the effect of pre-processing operations on setting-\romannumeral2 is more sensitive than that on setting-\romannumeral1, compared to the baseline model on KiTS19~\cite{kutikov2009renal}. Specifically, 1) \textbf{larger patching sizes result in better overall model performance}. When the patching size is relatively small (\eg, 32), the segmentation model is corrupted on both KiTS19~\cite{kutikov2009renal} and LiTS~\cite{bilic2019liver}. When the patching size is set to 192, the model on KiTS19~\cite{kutikov2009renal} achieves the best performance on setting-\romannumeral2. Experimental results on LiTS~\cite{bilic2019liver} show that the model performance gain increases gradually with the increase of patching size. Although there is a slight performance drop on setting-\romannumeral1 under Patching$_{192}$ compared to the baseline performance, this arises from the experimental setting on the image scale rather than the patching scheme~\cite{isensee2021nnu}. This experimental phenomenon is consistent with the conclusions of previous papers in~\cite{isensee2021nnu,yeung2022unified}, which recommends using as large an image patch size as GPU memory can accommodate. 
2) \textbf{OverSam~\cite{krawczyk2019radial,he2008adasyn} improves model performance overall}. In particular, + OverSam on KiTS19~\cite{kutikov2009renal} brings a remarkable performance gain of $0.84\%$ Recall, $0.43\%$ Dice, and $0.73\%$ IoU on setting-\romannumeral1 and $9.11\%$ Recall, $6.98\%$ Precision, $6.06\%$ Dice, and $5.05\%$ IoU on setting-\romannumeral2. 
3) \textbf{Without ReSam~\cite{ulyanov2016instance}, model performance significantly reduces on both KiTS19~\cite{kutikov2009renal} and LiTS~\cite{bilic2019liver}}. For example, on KiTS19, the model can result in a performance decrease of $90.83\%$ Recall, $69.71\%$ Precision, $92.15\%$ Dice, and $87.17\%$ IoU on setting-\romannumeral1. Surprisingly, the model is completely corrupted without using ReSam under setting-\romannumeral2. 
4) \textbf{IntesNorm~\cite{shah2011evaluating} has a relatively weak effect on results on both setting-\romannumeral1 and setting-\romannumeral2 on KiTS19.} For example, without using IntesNorm, the baseline 3D-UNet~\cite{cciccek20163d} only reduces the max model performance of $0.78\%$ Recall, $0.30\%$ Dice, and $0.70\%$ IoU on setting-\romannumeral1 and $0.20\%$ Recall and $0.00\%$ Dice on setting-\romannumeral2, respectively. However, IntesNorm has a significant effect on experimental results on LiTS~\cite{bilic2019liver}. This phenomenon shows that the same method has different effects on different datasets regarding data pre-processing strategies. The experimental results from 2) to 4) validate the importance of OverSam, ReSam, and IntesNorm in MedISeg.
\subsection{Data Augmentation}
\label{sec:3.3}
Data augmentation is used to address problems of insufficient training samples and overfitting~\cite{shorten2019survey,taylor2018improving,hataya2022meta,wang2018generative,xu2022comprehensive}. In particular, for medical images, data augmentation is frequently employed~\cite{zhao2019data,xu2020automatic}. The data augmentation schemes used in MedISeg can be broadly classified into two categories: geometric transformation-based data augmentation (GTAug)~\cite{taylor2018improving} and generative adversarial network (GAN)-based data augmentation (GANAug)~\cite{wang2018generative}.
\begin{table*}[t]
\begin{center}
\renewcommand\arraystretch{1.5}
\setlength{\tabcolsep}{5pt}{
\caption{Experimental results on data augmentation schemes. ``GTAug'' and ``GANAug'' denote the geometric transformation-based and the GAN-based data augmentation~\cite{isola2017image}, respectively. ``All'' denotes that both GTAug and GANAug are used in the model training phase.}
\vspace{-3mm}
\begin{tabular}{ r | c c c c | c c c c} 
 Methods & Recall (\%) & Percision (\%) & Dice (\%) & IoU (\%) & Recall (\%) & Percision (\%) & Dice (\%) & IoU (\%) \\ 
\hline \hline 
\multirow{2}{*}{2D-UNet~\cite{ronneberger2015u}} & \multicolumn{4}{c}{ISIC 2018~\cite{codella2019skin}} & \multicolumn{4}{c}{CoNIC~\cite{graham2021lizard}} \\
\cline{2-9}
~ & 88.18 & 89.88 & 86.89 & 85.80 & 78.12 & 77.25 & 77.23 & 77.58 \\
\cdashline{1-9}[0.8pt/2pt]
+ GTAug-A & 87.67$_{\color{blue}{-0.51}}$ & 90.19$_{\color{red}{+0.31}}$ & 86.87$_{\color{blue}{-0.02}}$ & 85.68$_{\color{blue}{-0.12}}$ & 78.94$_{\color{red}{+0.82}}$ & 77.54$_{\color{red}{+0.29}}$ & 77.87$_{\color{red}{+0.64}}$ & 78.05$_{\color{red}{+0.47}}$  \\
+ GTAug-B & 88.32$_{\color{red}{+0.14}}$ & 91.11$_{\color{red}{+1.23}}$ & 88.07$_{\color{red}{+1.18}}$ & 86.98$_{\color{red}{+1.18}}$ & 79.28$_{\color{red}{+1.16}}$ & 82.53$_{\color{red}{+5.28}}$ & 80.33$_{\color{red}{+3.10}}$ & 80.35$_{\color{red}{+2.77}}$ \\
+ GANAug~\cite{isola2017image} & 87.78$_{\color{blue}{-0.40}}$ & 89.59$_{\color{blue}{-0.29}}$ & 86.47$_{\color{blue}{-0.42}}$ & 85.64$_{\color{blue}{-0.16}}$ & 78.43$_{\color{red}{+0.31}}$ & 77.05$_{\color{blue}{-0.20}}$ & 77.37$_{\color{red}{+0.14}}$ & 77.62$_{\color{red}{+0.04}}$ \\
+ All & 87.77$_{\color{blue}{-0.41}}$ & 90.25$_{\color{red}{+0.37}}$ & 87.30$_{\color{red}{+0.41}}$ & 86.63$_{\color{red}{+0.83}}$ & 81.17$_{\color{red}{+3.05}}$ & 80.27$_{\color{red}{+3.02}}$ & 80.39$_{\color{red}{+3.16}}$ & 80.13$_{\color{red}{+2.55}}$ \\
\hline \hline
\multirow{2}{*}{3D-UNet~\cite{cciccek20163d}} & \multicolumn{4}{c}{KiTS19~\cite{heller2021state}:~settings-\romannumeral1} & \multicolumn{4}{c}{KiTS19~\cite{heller2021state}:~settings-\romannumeral2} \\
\cline{2-9}
~ & 91.01 & 95.20 & 92.50 & 87.35 & 27.35 & 46.71 & 29.63 & 21.51 \\
\cdashline{1-9}[0.8pt/2pt]
{+ GTAug-A} & 89.60$_{\color{blue}{-1.41}}$ & 95.44$_{\color{red}{+0.24}}$ & 91.65$_{\color{blue}{-0.85}}$ & 86.13$_{\color{blue}{-1.22}}$ & 21.81$_{\color{blue}{-5.54}}$ & 45.89$_{\color{blue}{-0.82}}$ & 25.68$_{\color{blue}{-3.95}}$ & 18.19$_{\color{blue}{-3.32}}$  \\
{+ GTAug-B} & 84.40$_{\color{blue}{-6.61}}$ & 94.58$_{\color{blue}{-0.62}}$ & 88.00$_{\color{blue}{-4.50}}$ & 80.94$_{\color{blue}{-6.41}}$ & 14.02$_{\color{blue}{-13.33}}$ & 37.70$_{\color{blue}{-9.01}}$ & 15.64$_{\color{blue}{-13.99}}$ & 10.71$_{\color{blue}{-10.80}}$  \\
{+ GANAug~\cite{isola2017image}} & 91.89$_{\color{red}{+0.88}}$ & 94.88$_{\color{blue}{-0.32}}$ & 92.87$_{\color{red}{+0.37}}$ & 87.98$_{\color{red}{+0.63}}$ & 29.19$_{\color{red}{+1.84}}$ & 47.69$_{\color{red}{+0.98}}$ & 30.99$_{\color{red}{+1.36}}$ & 22.89$_{\color{red}{+1.38}}$  \\
{+ All} & 85.97$_{\color{blue}{-5.04}}$ & 90.55$_{\color{blue}{-4.65}}$ & 86.93$_{\color{blue}{-5.57}}$ & 78.91$_{\color{blue}{-8.44}}$ & 6.86$_{\color{blue}{-20.49}}$ & 30.39$_{\color{blue}{-16.32}}$ & 9.93$_{\color{blue}{-19.70}}$ & 6.29$_{\color{blue}{-15.22}}$
\end{tabular}
\label{tab5}}
\vspace{-4mm}
\end{center}
\end{table*}
\begin{table}[t]
\begin{center}
\renewcommand\arraystretch{1.5}
\setlength{\tabcolsep}{1.5pt}{
\caption{Experimental results on 3D LiTS dataset~\cite{bilic2019liver} on different data augmentation schemes. ``GTAug'' and ``GANAug'' denote the geometric transformation-based and the generative adversarial network-based data augmentation~\cite{isola2017image}, respectively. ``All'' denotes that both GTAug and GANAug are used in the model training phase.}
\vspace{-2mm}
\begin{tabular}{ r | c c c c} 
 Methods & Recall (\%) & Percision (\%) & Dice (\%) & IoU (\%) \\ 
\hline \hline 
3D-UNet~\cite{cciccek20163d}  & 89.33 & 84.03 & 86.11 & 76.44 \\ 
\cdashline{1-5}[0.8pt/2pt]
{+ GTAug-A} & 90.28$_{\color{red}{+0.95}}$ & 84.24$_{\color{red}{+0.21}}$ & 86.62$_{\color{red}{+0.51}}$ & 76.89$_{\color{red}{+0.45}}$  \\
{+ GTAug-B} & 84.85$_{\color{blue}{-4.48}}$ & 81.60$_{\color{blue}{-2.43}}$ & 82.45$_{\color{blue}{-3.66}}$ & 70.97$_{\color{blue}{-5.47}}$  \\
{+ GANAug~\cite{isola2017image}} & 75.83$_{\color{blue}{-13.50}}$ & 76.42$_{\color{blue}{-7.61}}$ & 75.00$_{\color{blue}{-11.11}}$ & 61.49$_{\color{blue}{-14.95}}$ \\
{+ All} & 70.32$_{\color{blue}{-19.01}}$ & 77.04$_{\color{blue}{-6.99}}$ & 71.69$_{\color{blue}{-14.42}}$ & 57.89$_{\color{blue}{-18.55}}$ 
\end{tabular}
\label{tab6}}
\vspace{-3mm}
\end{center}
\end{table}

\subsubsection{GTAug} To eliminate the impact of geometric object variations in training images, such as positions, scales, and viewing angles, we employ GTAug~\cite{taylor2018improving}. GTAug is a widely used data augmentation method that includes flipping, cropping, rotations, translating, color jittering, contrast, simulation of low resolution, Gaussian noise injection, mixing images, random erasing, Gaussian blur, mixup, and cutmix~\cite{shorten2019survey,xu2022comprehensive,yeung2022unified,hataya2022meta}. In our work, we select several commonly used data augmentation methods, including random brightness contrast (with brightness limit = $0.2$, contrast limit = $0.2$, and $p = 0.5$), random gamma (with gamma limit = $(80, 120)$, and $p = 0.5$), CLAHE, random noise with $p = 0.5$, gamma adjust with $p = 0.5$, shift scale rotate (with shift limit=$0.1$, scale limit=$0.1$, rotate limit=$45$, and $p=0.5$), horizontal flip with $p = 0.5$, vertical flip with $p = 0.5$, random scale([0.85, 1.25]) with $p = 0.5$, random rotation([90, 180, 279]), and random mirroring along the X, Y, Z axial directions in a MedISeg model.

We divide the above data augmentation schemes into two groups and implement two types of data augmentation, named GTAug-A (\ie, pixel-level transform) and GTAug-B (\ie, spatial-level transform). Specifically, for experiments on 2D-UNet~\cite{ronneberger2015u}, we use random brightness contrast, random gamma, and CLAHE in GTAug-A, and shift scale rotate, horizontal flip, and vertical flip in GTAug-B. All schemes are deployed with their default parameters. For experiments on 3D-UNet~\cite{cciccek20163d}, we use random noise and gamma adjust in GTAug-A, and random scale, random rotation, and random mirroring in GTAug-B, all with their default parameters.

\subsubsection{GANAug}
The intrinsic prerequisite for data augmentation is to introduce the domain knowledge or other incremental information into the training dataset~\cite{tran2021data,zhao2020differentiable,karras2020training}. From this aspect, GANAug can be viewed as a loss function that focuses on guiding the network to generate some real data that is close to the source dataset domain~\cite{shorten2019survey,shin2018medical}. Especially, given the small dataset, the data distribution fitted by the generative model of GAN is better than that of the discriminative model. Therefore, GANAug is suitable for MedISeg task~\cite{yi2019generative,nie2017medical}. In our work, the classical pixel-level image-to-image translation with conditional adversarial networks~\cite{isola2017image} is used with its default setting. 

\subsubsection{Experimental Results}
In Tables~\ref{tab5} and \ref{tab6}, we present the experimental results of different data augmentation schemes. We observe that, in comparison with the baseline 2D-UNet~\cite{ronneberger2015u} on ISIC 2018~\cite{codella2019skin} in Table~\ref{tab5}, GTAug-B significantly improves the recognition performance according to all evaluation metrics, while GTAug-A slightly reduces the performance under three metrics, namely $-0.51\%$ Recall, $-0.02\%$ Dice, and $-0.12\%$ IoU. These observations confirm that \textbf{the spatial-level transform outperforms the pixel-level transform on 2D images.} In contrast to GTAug, GANAug achieves lower performance, with $-0.40\%$ Recall, $-0.29\%$ Precision, $-0.42\%$ Dice, and $-0.16\%$ IoU reductions. The overall performance of GANAug is worse than that of GTAug. When both GTAug and GANAug are applied to 2D-UNet~\cite{ronneberger2015u} on ISIC 2018~\cite{codella2019skin} (+ All in Table~\ref{tab5}), the model's performance improves according to three evaluation metrics, namely $0.37\%$ Precision, $0.41\%$ Dice, and $0.83\%$ IoU, resulting in a final performance of $87.77\%$ Recall, $90.25\%$ Precision, $87.30\%$ Dice, and $86.63\%$ IoU. Comparing the results on the baseline 2D-UNet~\cite{ronneberger2015u} on CoNIC~\cite{graham2021conic} and ISIC 2018~\cite{codella2019skin}, we observe that almost all values under the evaluation metrics are improved. The comparison of the experimental results on ISIC 2018~\cite{codella2019skin} and CoNIC~\cite{graham2021conic} shows that GTAug-B can achieve better results on both datasets than GTAug-A and GANAug, highlighting the importance of shift scale rotate, horizontal flip, and vertical flip. Furthermore, the above two sets of experimental results indicate that \textbf{we should choose the appropriate data augmentation methods when faced with different datasets}, as discussed in Section~\ref{sec:3.1}.

Regarding the experimental results on 3D KiTS19~\cite{kutikov2009renal} and LiTS~\cite{bilic2019liver}, we find that, in general, \textbf{GANAug can boost the maximum performance gain, while GTAug is barely beneficial on KiTS19~\cite{kutikov2009renal}}. In contrast, the experimental results on LiTS~\cite{bilic2019liver} show that only GTAug-A is beneficial, while GTAug-B and GANAug are detrimental to the performance. For instance, using GTAug on 3D-UNet on KiTS19~\cite{kutikov2009renal} reduces the maximum performance by $-13.33\%$, $9.01\%$, $13.99\%$, and $10.80\%$ for Recall, Precision, Dice, and IoU, respectively. In comparison, GANAug achieves performance gains of $0.88\%$ Recall, $0.37\%$ Dice, and $0.63\%$ IoU in setting-$\textrm{\romannumeral1}$ and $1.84\%$ Recall, $0.98\%$ Precision, $1.36\%$ Dice, and $1.38\%$ IoU in setting-$\textrm{\romannumeral2}$. When both GTAug and GANAug are applied to KiTS19~\cite{kutikov2009renal}, the model's performance is even lower, with $85.97\%$ Recall, $90.55\%$ Precision, $86.93\%$ Dice, and $78.91\%$ IoU in setting-$\textrm{\romannumeral1}$ and $6.86\%$ Recall, $30.39\%$ Precision, $9.93\%$ Dice, and $6.29\%$ IoU in setting-$\textrm{\romannumeral2}$. In particular, the results in setting-$\textrm{\romannumeral2}$ show the maximum performance reduction of $-20.49\%$, $-16.32\%$, $-19.70\%$, and $-15.22\%$ for Recall, Precision, Dice, and IoU, respectively. The experimental results on LiTS~\cite{bilic2019liver} in Table~\ref{tab6} show that GTAug-A can improve every evaluation metric by $0.95\%$ Recall, $0.21\%$ Precision, $0.51\%$ Dice, and $0.45\%$ IoU. In contrast, implementing both GANAug and GTAug at the same time (\ie, +All) leads to the most performance degradation. \textbf{These experimental results on KiTS19~\cite{kutikov2009renal} and LiTS~\cite{bilic2019liver} demonstrate the importance of selecting the appropriate data augmentation methods based on the specific dataset conditions}.
\subsection{Model Implementation}
\label{sec:3.4}
MedISeg models typically consist of many implementation details, which is particularly true in current research. In our work, we investigate the effectiveness of three commonly used implementation techniques categorized as: deep supervision (DeepS)~\cite{lee2015deeply}; class balance loss (CBL)~\cite{zang2021fasa}, which includes four loss functions (\ie, CBL$_{\textrm{Dice}}$~\cite{crum2006generalized}, CBL$_{\textrm{Focal}}$~\cite{lin2017focal}, CBL$_{\textrm{Tvers}}$~\cite{salehi2017tversky}, and CBL$_{\textrm{WCE}}$~\cite{phan2020resolving}); online hard example mining (OHEM)~\cite{shrivastava2016training}; and instance normalization (IntNorm)~\cite{ioffe2015batch}. Besides, as Transformer-based methods have gained increasing attention for MedISeg in recent years, we also present experimental results on ViT~\cite{dosovitskiy2020image}, MAE~\cite{he2022masked}, and MoCo~\cite{chen2021empirical} for comparison in this section.

\begin{table*}[t]
\begin{center}
\renewcommand\arraystretch{1.5}
\setlength{\tabcolsep}{5pt}{
\caption{Experimental results on some model implementation tricks. ``DeepS'', ``CBL'', ``OHEM'', and IntNorm denotes deep supervision~\cite{lee2015deeply}, class balance loss~\cite{zang2021fasa}, online hard example mining~\cite{shrivastava2016training}, and instance normalization~\cite{ioffe2015batch}, respectively.}
\vspace{-2mm}
\begin{tabular}{ r | c c c c | c c c c} 
Methods & Recall (\%) & Percision (\%) & Dice (\%) & IoU (\%) & Recall (\%) & Percision (\%) & Dice (\%) & IoU (\%) \\
\hline \hline 
\multirow{2}{*}{2D-UNet~\cite{ronneberger2015u}} & \multicolumn{4}{c}{ISIC 2018~\cite{codella2019skin}} & \multicolumn{4}{c}{CoNIC~\cite{graham2021lizard}} \\
\cline{2-9}
~ & 88.18 & 89.88 & 86.89 & 85.80 & 78.12 & 77.25 & 77.23 & 77.58 \\
\cdashline{1-9}[0.8pt/2pt]
+ DeepS~\cite{lee2015deeply} & 88.91$_{\color{red}{+0.73}}$ & 89.85$_{\color{blue}{-0.03}}$ & 87.42$_{\color{red}{+0.53}}$ & 86.18$_{\color{red}{+0.38}}$ & 78.16$_{\color{red}{+0.04}}$ & 77.02$_{\color{blue}{-0.23}}$ & 77.13$_{\color{blue}{-0.10}}$ & 77.46$_{\color{blue}{-0.12}}$  \\
\cdashline{1-9}[0.8pt/2pt]
+ CBL$_{\textrm{Dice}}$~\cite{crum2006generalized} & 89.59$_{\color{red}{+1.41}}$ & 89.89$_{\color{red}{+0.01}}$ & 87.71$_{\color{red}{+0.82}}$ & 86.36$_{\color{red}{+0.56}}$ & 79.38$_{\color{red}{+1.26}}$ & 78.37$_{\color{red}{+1.12}}$ & 78.36$_{\color{red}{+1.13}}$ & 78.51$_{\color{red}{+0.93}}$  \\
+ CBL$_{\textrm{Focal}}$~\cite{lin2017focal} & 88.06$_{\color{blue}{-0.12}}$ & 87.32$_{\color{blue}{-2.56}}$ & 85.41$_{\color{blue}{-1.48}}$ & 84.65$_{\color{blue}{-1.15}}$ & 81.78$_{\color{red}{+3.66}}$ & 73.75$_{\color{blue}{-3.50}}$ & 77.12$_{\color{blue}{-0.11}}$ & 77.24$_{\color{blue}{-0.34}}$ \\
+ CBL$_{\textrm{Tvers}}$~\cite{salehi2017tversky} & 89.40$_{\color{red}{+1.22}}$ & 90.19$_{\color{red}{+0.31}}$ & 87.87$_{\color{red}{+0.98}}$ & 86.42$_{\color{red}{+0.62}}$ & 78.65$_{\color{red}{+0.53}}$ & 78.79$_{\color{red}{+1.54}}$ & 78.23$_{\color{red}{+1.00}}$ & 78.40$_{\color{red}{+0.82}}$ \\
+ CBL$_{\textrm{WCE}}$~\cite{phan2020resolving} & 89.72$_{\color{red}{+1.54}}$ & 88.15$_{\color{blue}{-1.73}}$ & 86.72$_{\color{blue}{-0.17}}$ & 85.63$_{\color{blue}{-0.17}}$ & 82.24$_{\color{red}{+4.12}}$ & 74.08$_{\color{blue}{-3.17}}$ & 77.54$_{\color{red}{+0.31}}$ & 77.58$_{\color{red}{+0.00}}$  \\
\cdashline{1-9}[0.8pt/2pt]
+ OHEM~\cite{shrivastava2016training} & 88.06$_{\color{blue}{-0.12}}$ & 89.81$_{\color{blue}{-0.07}}$ & 86.85$_{\color{blue}{-0.04}}$ & 85.80$_{\color{red}{+0.00}}$ & 77.35$_{\color{blue}{-0.77}}$ & 77.72$_{\color{red}{+0.47}}$ & 77.06$_{\color{blue}{-0.17}}$ & 77.47$_{\color{blue}{-0.11}}$ \\
\hline \hline
\multirow{2}{*}{3D-UNet~\cite{cciccek20163d}} & \multicolumn{4}{c}{KiTS19~\cite{heller2021state}:~settings-\romannumeral1} & \multicolumn{4}{c}{KiTS19~\cite{heller2021state}:~settings-\romannumeral2} \\
\cline{2-9}
~ & 91.01 & 95.20 & 92.50 & 87.35 & 27.35 & 46.71 & 29.63 & 21.51  \\
\cdashline{1-9}[0.8pt/2pt]
{+ DeepS~\cite{lee2015deeply}} & 90.13$_{\color{blue}{-0.88}}$ & 95.07$_{\color{blue}{-0.13}}$ & 91.88$_{\color{blue}{-0.62}}$ & 86.42$_{\color{blue}{-0.93}}$ & 26.69$_{\color{blue}{-0.66}}$ & 46.03$_{\color{blue}{-0.68}}$ & 29.27$_{\color{blue}{-0.36}}$ & 21.73$_{\color{red}{+0.22}}$ \\
\cdashline{1-9}[0.8pt/2pt]
{+ CBL$_{\textrm{Dice}}$~\cite{crum2006generalized}} & 90.50$_{\color{blue}{-0.51}}$ & 93.24$_{\color{blue}{-1.96}}$ & 91.17$_{\color{blue}{-1.33}}$ & 85.07$_{\color{blue}{-2.28}}$ & 41.91$_{\color{red}{+14.56}}$ & 44.70$_{\color{blue}{-2.01}}$ & 37.15$_{\color{red}{+7.52}}$ & 26.74$_{\color{red}{+5.23}}$ \\
{+ CBL$_{\textrm{Focal}}$~\cite{lin2017focal}} & 74.88$_{\color{blue}{-16.13}}$ & 95.64$_{\color{red}{+0.44}}$ & 81.29$_{\color{blue}{-11.21}}$ & 72.27$_{\color{blue}{-15.08}}$ & 9.72$_{\color{blue}{-17.63}}$ & 37.80$_{\color{blue}{-8.91}}$ & 12.98$_{\color{blue}{-16.65}}$ & 8.29$_{\color{blue}{-13.22}}$  \\
{+ CBL$_{\textrm{Tvers}}$~\cite{salehi2017tversky}} & 87.72$_{\color{blue}{-3.29}}$ & 93.94$_{\color{blue}{-1.26}}$ & 89.99$_{\color{blue}{-2.51}}$ & 83.13$_{\color{blue}{-4.22}}$ & 35.89$_{\color{red}{+8.54}}$ & 44.15$_{\color{blue}{-2.56}}$ & 33.49$_{\color{red}{+3.86}}$ & 23.68$_{\color{red}{+2.17}}$  \\
{+ CBL$_{\textrm{WCE}}$~\cite{phan2020resolving}} & 92.57$_{\color{red}{+1.56}}$ & 88.01$_{\color{blue}{-7.19}}$ & 89.49$_{\color{blue}{-3.01}}$ & 82.49$_{\color{blue}{-4.86}}$ & 36.07$_{\color{red}{+8.72}}$ & 33.53$_{\color{blue}{-13.18}}$ & 29.85$_{\color{red}{+0.22}}$ & 21.39$_{\color{blue}{-0.12}}$  \\
\cdashline{1-9}[0.8pt/2pt]
{+ OHEM~\cite{shrivastava2016training}} & 91.01$_{\color{red}{+0.00}}$ & 94.81$_{\color{blue}{-0.39}}$ & 92.17$_{\color{blue}{-0.33}}$ & 86.94$_{\color{blue}{-0.41}}$ & 28.45$_{\color{red}{+1.10}}$ & 46.68$_{\color{blue}{-0.03}}$ & 30.58$_{\color{red}{+0.95}}$ & 22.21$_{\color{red}{+0.70}}$  \\
{+ IntNorm~\cite{ioffe2015batch}} & 73.53$_{\color{blue}{-17.48}}$ & 85.64$_{\color{blue}{-9.56}}$ & 77.45$_{\color{blue}{-15.05}}$ & 65.93$_{\color{blue}{-21.42}}$ & 16.92$_{\color{blue}{-10.43}}$ & 32.84$_{\color{blue}{-13.87}}$ & 18.45$_{\color{blue}{-11.18}}$ & 11.67$_{\color{blue}{-9.84}}$
\end{tabular}
\label{tab7}}
\vspace{-4mm}
\end{center}
\end{table*}
\begin{table}[t]
\begin{center}
\renewcommand\arraystretch{1.5}
\setlength{\tabcolsep}{1.5pt}{
\caption{Experimental results on 3D LiTS dataset~\cite{bilic2019liver} on some model implementation tricks. ``DeepS'', ``CBL'', ``OHEM'', and IntNorm denotes deep supervision~\cite{lee2015deeply}, class balance loss~\cite{zang2021fasa}, online hard example mining~\cite{shrivastava2016training}, and instance normalization~\cite{ioffe2015batch}, respectively.}
\begin{tabular}{ r | c c c c} 
 Methods & Recall (\%) & Percision (\%) & Dice (\%) & IoU (\%) \\ 
\hline \hline 
3D-UNet~\cite{cciccek20163d} & 89.33 & 84.03 & 86.11 & 76.44 \\ 
\cdashline{1-5}[0.8pt/2pt]
{+ DeepS~\cite{lee2015deeply}} & 90.42$_{\color{red}{+1.09}}$ & 84.02$_{\color{blue}{-0.01}}$ & 86.60$_{\color{red}{+0.49}}$ & 77.17$_{\color{red}{+0.73}}$  \\
\cdashline{1-5}[0.8pt/2pt]
{+ CBL$_{\textrm{Dice}}$~\cite{crum2006generalized}} & 83.09$_{\color{blue}{-6.24}}$ & 71.34$_{\color{blue}{-12.69}}$ & 75.51$_{\color{blue}{-10.60}}$ & 62.45$_{\color{blue}{-13.99}}$ \\
{+ CBL$_{\textrm{Focal}}$~\cite{lin2017focal}} & 88.37$_{\color{blue}{-0.96}}$ & 81.61$_{\color{blue}{-2.42}}$ & 84.16$_{\color{blue}{-1.95}}$ & 73.16$_{\color{blue}{-3.28}}$ \\
{+ CBL$_{\textrm{Tvers}}$~\cite{salehi2017tversky}} & 86.60$_{\color{blue}{-2.73}}$ & 72.93$_{\color{blue}{-11.10}}$ & 78.70$_{\color{blue}{-7.41}}$ & 65.52$_{\color{blue}{-10.92}}$  \\
{+ CBL$_{\textrm{WCE}}$~\cite{phan2020resolving}} & 91.18$_{\color{red}{+1.85}}$ & 80.27$_{\color{blue}{-3.76}}$ & 84.92$_{\color{blue}{-1.19}}$ & 74.57$_{\color{blue}{-1.87}}$ \\
\cdashline{1-5}[0.8pt/2pt]
{+ OHEM~\cite{shrivastava2016training}} & 90.14$_{\color{red}{+0.81}}$ & 85.64$_{\color{red}{+1.61}}$ & 87.35$_{\color{red}{+1.24}}$ & 78.24$_{\color{red}{+1.80}}$  \\
{+ IntNorm~\cite{ioffe2015batch}} & 77.23$_{\color{blue}{-12.10}}$ & 87.27$_{\color{red}{+3.24}}$ & 80.94$_{\color{blue}{-5.17}}$ & 68.62$_{\color{blue}{-7.82}}$ 
\end{tabular}
\label{tab8}}
\vspace{-8mm}
\end{center}
\end{table}
\begin{table*}[t]
\begin{center}
\renewcommand\arraystretch{1.5}
\setlength{\tabcolsep}{5pt}{
\caption{Experimental results on Transformer-based methods, where ``/16'' and ``/32'' are the number of heads in the multi-head attention mechanism.}
\vspace{-2mm}
\begin{tabular}{ r | c c c c | c c c c} 
Methods & Recall (\%) & Percision (\%) & Dice (\%) & IoU (\%) & Recall (\%) & Percision (\%) & Dice (\%) & IoU (\%) \\
\hline \hline 
\multirow{2}{*}{2D-UNet~\cite{ronneberger2015u}} & \multicolumn{4}{c}{ISIC 2018~\cite{codella2019skin}} & \multicolumn{4}{c}{CoNIC~\cite{graham2021lizard}} \\
\cline{2-9}
~ & 88.18 & 89.88 & 86.89 & 85.80 & 78.12 & 77.25 & 77.23 & 77.58 \\
\cdashline{1-9}[0.8pt/2pt]
ViT-B/16~\cite{dosovitskiy2020image} & 87.56$_{\color{blue}{-0.62}}$ & 87.84$_{\color{blue}{-2.04}}$ & 85.22$_{\color{blue}{-1.67}}$ & 84.45$_{\color{blue}{-1.35}}$ & 51.83$_{\color{blue}{-26.29}}$ & 63.27$_{\color{blue}{-13.98}}$ & 55.57$_{\color{blue}{-21.66}}$ & 67.10$_{\color{blue}{-10.48}}$ \\
ViT-B/16-MAE~\cite{he2022masked} & 87.28$_{\color{blue}{-0.90}}$ & 87.62$_{\color{blue}{-2.26}}$ & 84.83$_{\color{blue}{-2.06}}$ & 84.21$_{\color{blue}{-1.59}}$ & 51.62$_{\color{blue}{-26.50}}$ & 62.98$_{\color{blue}{-14.27}}$ & 55.24$_{\color{blue}{-21.99}}$ & 66.45$_{\color{blue}{-11.13}}$ \\
ViT-B/16-MoCo v3~\cite{chen2021empirical} & 87.93$_{\color{blue}{-0.25}}$ & 87.67$_{\color{blue}{-2.21}}$ & 85.34$_{\color{blue}{-1.55}}$ & 84.58$_{\color{blue}{-1.22}}$ & 52.13$_{\color{blue}{-25.99}}$ & 63.35$_{\color{blue}{-13.90}}$ & 55.78$_{\color{blue}{-21.45}}$ & 67.22$_{\color{blue}{-10.36}}$ \\
ViT-L/16~\cite{dosovitskiy2020image} & 87.66$_{\color{blue}{-0.52}}$ & 87.53$_{\color{blue}{-2.35}}$ & 85.04$_{\color{blue}{-1.85}}$ & 84.31$_{\color{blue}{-1.49}}$ & 51.89$_{\color{blue}{-26.23}}$ & 63.10$_{\color{blue}{-14.15}}$ & 55.71$_{\color{blue}{-21.52}}$ & 67.01$_{\color{blue}{-10.57}}$ \\
\cdashline{1-9}[0.8pt/2pt]
ViT-B/32~\cite{dosovitskiy2020image} & 83.61$_{\color{blue}{-4.57}}$ & 83.17$_{\color{blue}{-6.71}}$ & 80.99$_{\color{blue}{-5.90}}$ & 81.51$_{\color{blue}{-4.29}}$ & 48.96$_{\color{blue}{-29.16}}$ & 60.65$_{\color{blue}{-16.60}}$ & 52.63$_{\color{blue}{-24.60}}$ & 63.88$_{\color{blue}{-13.70}}$ \\
ViT-B/32-MAE~\cite{he2022masked} & 84.21$_{\color{blue}{-3.97}}$ & 82.96$_{\color{blue}{-6.92}}$ & 81.20$_{\color{blue}{-5.69}}$ & 81.60$_{\color{blue}{-4.20}}$ & 49.51$_{\color{blue}{-28.61}}$ & 60.23$_{\color{blue}{-17.02}}$ & 52.90$_{\color{blue}{-24.33}}$ & 63.94$_{\color{blue}{-13.64}}$ \\
ViT-B/32-MoCo v3~\cite{chen2021empirical} & 83.97$_{\color{blue}{-4.21}}$ & 82.78$_{\color{blue}{-7.10}}$ & 80.88$_{\color{blue}{-6.01}}$ & 81.39$_{\color{blue}{-4.41}}$ & 49.36$_{\color{blue}{-28.76}}$ & 59.95$_{\color{blue}{-17.30}}$ & 52.76$_{\color{blue}{-24.47}}$ & 63.82$_{\color{blue}{-13.76}}$ \\
ViT-L/32~\cite{dosovitskiy2020image} & 84.37$_{\color{blue}{-3.81}}$ & 83.16$_{\color{blue}{-6.72}}$ & 81.32$_{\color{blue}{-5.57}}$ & 81.65$_{\color{blue}{-4.15}}$ & 49.32$_{\color{blue}{-28.80}}$ & 60.58$_{\color{blue}{-16.67}}$ & 53.03$_{\color{blue}{-24.20}}$ & 65.04$_{\color{blue}{-12.54}}$
\end{tabular}
\label{tabtrans}}
\vspace{-4mm}
\end{center}
\end{table*}
\subsubsection{DeepS}
DeepS~\cite{lee2015deeply} is an auxiliary learning technique that was initially proposed in DSN~\cite{lee2015deeply} and subsequently used for image classification~\cite{wang2015training}. This technique is aimed at supervising the backbone network by adding an auxiliary classifier or segmenter on some intermediate hidden layers in a direct or indirect manner~\cite{zhang2021self,phuong2019distillation,lin2017feature,zhang2020feature}. It can be used to address issues related to training gradient disappearance or slow convergence speed. For image segmentation, this technique is usually implemented by adding an image-level classification loss. In our work, we follow~\cite{phuong2019distillation,zhang2021self,yu2018learning} and extract feature maps from the last three decoder layers. We then use a $1 \times 1$ convolutional layer to project the lesion mask into the same channel size. Output feature maps from different layers are upsampled into the same spatial size as the input image for the segmentation head network~\cite{zhang2020causal,zhang2018context,zhang2020feature} by bilinear interpolations.

\subsubsection{CBL} CBL~\cite{zang2021fasa} is usually used to learn a general class weight, \ie, the weight for each class is only related to the object category. Compared to some traditional segmentation loss functions (\eg, cross-entropy loss) on the class imbalanced dataset~\cite{zang2021fasa,lee2015deeply}, CBL can improve model representational ability. 
In the used dataset, CBL introduces the effective number of samples to represent the expected volume representations of the selected dataset, and weights different classes by the number of effective samples rather than the number of original samples. In this paper, we mainly explore the effect of four commonly used CBL loss functions for the class imbalanced problem in medical image domain, including Dice loss (CBL$_{\textrm{Dice}}$)~\cite{crum2006generalized}, Focal loss (CBL$_{\textrm{Focal}}$)~\cite{lin2017focal}, Tversky loss (CBL$_{\textrm{Tvers}}$~\cite{salehi2017tversky}), and the weighted cross-entropy loss (CBL$_{\textrm{WCE}}$~\cite{phan2020resolving}). 

\subsubsection{OHEM~} The core idea of OHEM\cite{shrivastava2016training} is first to filter out some hard learning samples (\ie, images, objects, and pixels) via the loss function, and these selected hard examples all have a high impact on the recognition tasks~\cite{shrivastava2016training}. Then, these samples are applied to gradient descent in the model training process. Extensive experimental results on different vision tasks show that OHEM is not only efficient but also performs well on various datasets~\cite{chen2017deeplab,liu2019auto,chen2018encoder,zhao2017pyramid}. In our work, we validate the effectiveness of OHEM on both the 2D and 3D medical datasets with its default setting.

\subsubsection{IntNorm} IntNorm~\cite{ioffe2015batch} is a popular normalization algorithm that is suitable for recognition tasks with higher requirements on a single pixel. In its implementation, every single sample and all elements for a single channel of the sample are taken into consideration when computing the statistic normalization~\cite{ioffe2015batch,gatys2016image,ba2016layer}. In the medical image domain, an important reason why IntNorm is used is that the batch size is usually set to a small value (especially for 3D images) during the training process, which makes the use of batch normalization invalid. In this paper, we demonstrate the effect of the IntNorm by replacing it with the intensity normalization~\cite{isensee2021nnu} in 3D experiments. 

\subsubsection{Transformer-Based Methods.} We take Vit~\cite{dosovitskiy2020image}, MAE~\cite{he2022masked}, and MoCo~\cite{chen2021empirical} in our experiments. ViT divides the input image into multiple image patches, and then projects each patch into a fixed-length vector and sends it to the Transformer layer. The decoder gradually upsamples the obtained feature vector to the same size as the input image for prediction. Based on ViT, MAE~\cite{he2022masked} first masks random patches and tries to reconstruct them during training, which is a typical self-supervised learning method. MoCo~\cite{chen2021empirical} is also an unsupervised visual representation learning method which based on ViT. MoCo compares contrastive learning to the process of looking up a dictionary, and regards the dictionary as a queue and introduces the idea of momentum update.

\subsubsection{Experimental Results}
The experimental results of implementing detailed model implementation tricks on 2D-UNet~\cite{ronneberger2015u} for the 2D ISIC 2018~\cite{codella2019skin} and 2D CoNIC~\cite{graham2021conic} datasets and 3D-UNet~\cite{cciccek20163d} for the KiTS19~\cite{kutikov2009renal} and 3D LiTS~\cite{bilic2019liver} datasets are presented in Table~\ref{tab7} and Table~\ref{tab8}. The results indicate that \textbf{implementing these tricks leads to more significant accuracy improvements on 2D datasets than on 3D datasets.} For instance, implementing DeepS~\cite{lee2015deeply}, CBL$_{\textrm{Dice}}$~\cite{crum2006generalized}, and CBL$_{\textrm{Tvers}}$~\cite{salehi2017tversky} on 2D-UNet~\cite{ronneberger2015u} for ISIC 2018~\cite{codella2019skin} significantly boosts model performance. However, other tricks such as +CBL$_{\textrm{Focal}}$, +CBL$_{\textrm{WCE}}$, and +OHEM~\cite{shrivastava2016training} result in reduced model performance on most evaluation metrics for both 2D datasets. These observations and conclusions hold for 2D-UNet~\cite{ronneberger2015u} on both ISIC 2018~\cite{codella2019skin} and CoNIC~\cite{graham2021conic}. Notably, these tricks were initially designed for 2D datasets and validated on 2D datasets, which explains why they do not work well on 3D datasets.

Experimental results on 3D-UNet~\cite{cciccek20163d} for KiTS19~\cite{kutikov2009renal} and 3D LiTS~\cite{bilic2019liver} indicate that \textbf{implementing these model implementation tricks cannot significantly improve performance}. However, CBL$_{\textrm{Dice}}$ can bring performance gains of $14.56\%$ Recall, $7.52\%$ Dice, and $5.23\%$ IoU, and CBL$_{\textrm{Tvers}}$ can bring performance gains of $8.54\%$ Recall, $3.86\%$ Dice, and $2.17\%$ IoU on setting-$\textrm{\romannumeral2}$ for KiTS19~\cite{kutikov2009renal}. Additionally, OHEM~\cite{shrivastava2016training} can also improve model performance. The results in the last row of Table~\ref{tab7} show that the effect of IntNorm~\cite{ioffe2015batch} is not as good as that of intensity normalization~\cite{isensee2021nnu} on the 3D KiTS19 dataset. Replacing IntNorm with intensity normalization~\cite{isensee2021nnu} leads to a reduction in the maximum performance of $-17.48\%$ Recall, $-13.87\%$ Precision, $-15.05\%$ Dice, and $-21.42\%$ IoU. \textbf{The fact that 2D tricks do not work well on 3D datasets highlights the need to consider the dataset format in future model designs. To make these tricks work on 3D datasets, they may need to be revised before deployment in the application process.}

The experimental results presented in Table~\ref{tabtrans} show that the performance of Transformer-based methods on ISIC 2018~\cite{codella2019skin} and CoNIC~\cite{graham2021lizard} is significantly lower than that of CNN-based methods. This may be because Transformer-based models lack sufficient inductive bias, and the amount of data in the training set is insufficient. In medical images, the target positions are relatively consistent, making inductive bias crucial in MedISeg. Furthermore, Transformer-based models typically have more parameters than their CNN-based counterparts, requiring a larger amount of data to optimize them. \textbf{These findings emphasize the importance of considering the amount of data available or modifying existing Transformer models to achieve satisfactory results in MedISeg.}
\subsection{Model Inference}
\label{sec:3.5}
In our work, we mainly explore two kinds of commonly used inference tricks, namely, test time augmentation (TTA) and the model ensemble (Ensemble).
\begin{table*}[t]
\begin{center}
\renewcommand\arraystretch{1.5}
\setlength{\tabcolsep}{5pt}{
\caption{Experimental results on model inference tricks. ``All'' denotes that both TTA$_{\textrm{GTAug}}$ and the corresponding model ensemble strategy are used.}
\vspace{-3mm}
\begin{tabular}{ r | c c c c | c c c c} 
Methods & Recall (\%) & Percision (\%) & Dice (\%) & IoU (\%) & Recall (\%) & Percision (\%) & Dice (\%) & IoU (\%) \\ 
\hline \hline 
\multirow{2}{*}{2D-UNet~\cite{ronneberger2015u}} & \multicolumn{4}{c}{ISIC 2018~\cite{codella2019skin}} & \multicolumn{4}{c}{CoNIC~\cite{graham2021lizard}} \\
\cline{2-9}
~ & 88.18 & 89.88 & 86.89 & 85.80 & 78.12 & 77.25 & 77.23 & 77.58 \\
\cdashline{1-9}[0.8pt/2pt]
+ TTA$_{\textrm{baseline}}$ & 88.55$_{\color{red}{+0.37}}$ & 90.29$_{\color{red}{+0.41}}$ & 87.42$_{\color{red}{+0.53}}$ & 86.26$_{\color{red}{+0.46}}$  & 79.13$_{\color{red}{+1.01}}$ & 80.62$_{\color{red}{+3.37}}$ & 79.37$_{\color{red}{+2.14}}$ & 79.41$_{\color{red}{+1.83}}$ \\
+ TTA$_{\textrm{GTAug-A}}$ & 88.28$_{\color{red}{+0.10}}$ & 90.46$_{\color{red}{+0.58}}$ & 87.37$_{\color{red}{+0.48}}$ & 86.13$_{\color{red}{+0.33}}$ & 80.19$_{\color{red}{+2.07}}$ & 80.57$_{\color{red}{+3.32}}$ & 80.00$_{\color{red}{+2.77}}$ & 79.86$_{\color{red}{+2.28}}$   \\
+ TTA$_{\textrm{GTAug-B}}$ & 90.21$_{\color{red}{+2.03}}$ & 90.94$_{\color{red}{+1.06}}$ & 88.94$_{\color{red}{+2.05}}$ & 87.59$_{\color{red}{+1.79}}$ & 80.04$_{\color{red}{+1.92}}$ & 83.81$_{\color{red}{+6.56}}$ & 81.32$_{\color{red}{+4.09}}$ & 81.22$_{\color{red}{+3.64}}$  \\
\cdashline{1-9}[0.8pt/2pt]
+ EnsAvg~\cite{kamnitsas2017ensembles} & 91.08$_{\color{red}{+2.90}}$ & 89.50$_{\color{blue}{-0.38}}$ & 88.52$_{\color{red}{+1.63}}$ & 87.21$_{\color{red}{+1.41}}$ & 79.97$_{\color{red}{+1.85}}$ & 79.90$_{\color{red}{+2.65}}$ & 79.45$_{\color{red}{+2.22}}$ & 79.85$_{\color{red}{+2.27}}$ \\
+ EnsVot~\cite{kamnitsas2017ensembles} & 91.04$_{\color{red}{+2.86}}$ & 89.41$_{\color{blue}{-0.47}}$ & 88.46$_{\color{red}{+1.57}}$ & 87.15$_{\color{red}{+1.35}}$ & 80.22$_{\color{red}{+2.10}}$ & 79.42$_{\color{red}{+2.17}}$ & 79.34$_{\color{red}{+2.11}}$ & 79.73$_{\color{red}{+2.15}}$ \\
\cdashline{1-9}[0.8pt/2pt]
+ All (TTA-EnsAvg) & 88.64$_{\color{red}{+0.46}}$ & 90.96$_{\color{red}{+1.08}}$ & 87.90$_{\color{red}{+1.01}}$ & 86.72$_{\color{red}{+0.92}}$ & 79.98$_{\color{red}{+1.86}}$ & 80.60$_{\color{red}{+3.35}}$ & 79.83$_{\color{red}{+2.60}}$ & 80.15$_{\color{red}{+2.57}}$  \\
+ All (TTA-EnsVot) & 88.61$_{\color{red}{+0.43}}$ & 90.86$_{\color{red}{+0.98}}$ & 87.87$_{\color{red}{+0.98}}$ & 86.69$_{\color{red}{+0.89}}$ & 79.89$_{\color{red}{+1.77}}$ & 80.56$_{\color{red}{+3.31}}$ & 79.76$_{\color{red}{+2.53}}$ & 80.09$_{\color{red}{+2.51}}$ \\
\hline \hline
\multirow{2}{*}{3D-UNet~\cite{cciccek20163d}} & \multicolumn{4}{c}{KiTS19~\cite{heller2021state}:~settings-\romannumeral1} & \multicolumn{4}{c}{KiTS19~\cite{heller2021state}:~settings-\romannumeral2} \\
\cline{2-9}
~ & 91.01 & 95.20 & 92.50 & 87.35 & 27.35 & 46.71 & 29.63 & 21.51  \\
\cdashline{1-9}[0.8pt/2pt]
+ TTA$_{\textrm{baseline}}$ & 72.94$_{\color{blue}{-18.07}}$ & 97.82$_{\color{red}{+2.62}}$ & 81.38$_{\color{blue}{-11.12}}$ & 72.08$_{\color{blue}{-15.27}}$ & 19.70$_{\color{blue}{-7.65}}$ & 39.60$_{\color{blue}{-7.11}}$ & 21.81$_{\color{blue}{-7.82}}$ & 15.39$_{\color{blue}{-6.12}}$ \\
{+ TTA$_{\textrm{GTAug-A}}$} & 71.32$_{\color{blue}{-19.69}}$ & 97.81$_{\color{red}{+2.61}}$ & 80.22$_{\color{blue}{-12.28}}$ & 70.60$_{\color{blue}{-16.75}}$ & 17.20$_{\color{blue}{-10.15}}$ & 38.28$_{\color{blue}{-8.43}}$ & 19.16$_{\color{blue}{-10.47}}$ & 13.27$_{\color{blue}{-8.24}}$  \\
{+ TTA$_{\textrm{GTAug-B}}$} & 82.43$_{\color{blue}{-8.58}}$ & 94.91$_{\color{blue}{-0.29}}$ & 86.65$_{\color{blue}{-5.85}}$ & 79.24$_{\color{blue}{-8.11}}$ & 10.59$_{\color{blue}{-16.76}}$ & 34.69$_{\color{blue}{-12.02}}$ & 14.17$_{\color{blue}{-15.46}}$ & 9.52$_{\color{blue}{-11.99}}$  \\
\cdashline{1-9}[0.8pt/2pt]
{+ EnsAvg~\cite{kamnitsas2017ensembles}} & 93.00$_{\color{red}{+1.99}}$ & 96.69$_{\color{red}{+1.49}}$ & 94.39$_{\color{red}{+1.89}}$ & 90.02$_{\color{red}{+2.67}}$ & 27.09$_{\color{blue}{-0.26}}$ & 54.71$_{\color{red}{+8.00}}$ & 29.15$_{\color{blue}{-0.48}}$ & 21.26$_{\color{blue}{-0.25}}$  \\
{+ EnsVot~\cite{kamnitsas2017ensembles}} & 92.90$_{\color{red}{+1.89}}$ & 96.62$_{\color{red}{+1.42}}$ & 94.31$_{\color{red}{+1.81}}$ & 89.87$_{\color{red}{+2.52}}$ & 27.65$_{\color{red}{+0.30}}$ & 56.65$_{\color{red}{+9.94}}$ & 29.44$_{\color{blue}{-0.19}}$ & 21.40$_{\color{blue}{-0.11}}$  \\
\cdashline{1-9}[0.8pt/2pt]
{+ All (TTA-EnsAvg)} & 79.75$_{\color{blue}{-11.26}}$ & 98.54$_{\color{red}{+3.34}}$ & 86.86$_{\color{blue}{-5.64}}$ & 79.01$_{\color{blue}{-8.34}}$ & 24.49$_{\color{blue}{-2.86}}$ & 48.17$_{\color{red}{+1.46}}$ & 27.02$_{\color{blue}{-2.61}}$ & 19.56$_{\color{blue}{-1.95}}$  \\
{+ All (TTA-EnsVot)} & 79.03$_{\color{blue}{-11.98}}$ & 98.73$_{\color{red}{+3.53}}$ & 86.31$_{\color{blue}{-6.19}}$ & 78.28$_{\color{blue}{-9.07}}$ & 23.38$_{\color{blue}{-3.97}}$ & 50.05$_{\color{red}{+3.34}}$ & 26.42$_{\color{blue}{-3.21}}$ & 19.11$_{\color{blue}{-2.40}}$ \\
\end{tabular}
\label{tab9}}
\vspace{-4mm}
\end{center}
\end{table*}

\subsubsection{TTA} 
TTA can be used to improve performance without training, so it has the potential to be a plug-and-play.
At the same time, it can improve the ability of model calibration, which is beneficial for visual tasks~\cite{shanmugam2021better,moshkov2020test}. In this work, we follow the same image augmentation strategy as in subsection~\ref{sec:3.3} by three aspects: 1) implementing the TTA strategy on the baseline model (\ie, TTA$_{\textrm{baseline}}$). 2) implementing the test time augmentation GTAug-A on the baseline model under the same data augmentation strategy GTAug-A (\ie, TTA$_{\textrm{GTAug-A}}$). 3) implementing the test time augmentation GTAug-B on the baseline model under the same data augmentation strategy GTAug-B (\ie, TTA$_{\textrm{GTAug-B}}$).

\subsubsection{Ensemble} 
The commonly used model ensemble methods are voting, averaging, stacking, and non-cross-stacking (blending)~\cite{kamnitsas2017ensembles,li2018fully}. In our work, we first independently conduct 5 training sets with different random seeds (\ie, $2022-2026$) obtaining 5 different models in fold-validation. Then, we choose the commonly used voting and averaging as the model ensemble strategies~\cite{kamnitsas2017ensembles}, termed as EnsVot and EnsAvg, respectively. 
\begin{table}[t]
\begin{center}
\renewcommand\arraystretch{1.5}
\setlength{\tabcolsep}{1.5pt}{
\caption{Experimental results on 3D LiTS dataset~\cite{bilic2019liver} on model inference tricks. ``All'' denotes that both TTA$_{\textrm{GTAug}}$ and the corresponding model ensemble strategy are used.}
\begin{tabular}{ r | c c c c} 
 Methods & Recall (\%) & Percision (\%) & Dice (\%) & IoU (\%) \\ 
\hline \hline 
3D-UNet~\cite{cciccek20163d} & 89.33 & 84.03 & 86.11 & 76.44 \\  
\cdashline{1-5}[0.8pt/2pt]
+ TTA$_{\textrm{baseline}}$ & 81.03$_{\color{blue}{-8.30}}$ & 85.35$_{\color{red}{+1.32}}$ & 82.32$_{\color{blue}{-3.79}}$ & 70.81$_{\color{blue}{-5.63}}$ \\
{+ TTA$_{\textrm{GTAug-A}}$} & 82.32$_{\color{blue}{-7.01}}$ & 85.28$_{\color{red}{+1.25}}$ & 83.06$_{\color{blue}{-3.05}}$ & 71.45$_{\color{blue}{-4.99}}$ \\
{+ TTA$_{\textrm{GTAug-B}}$} & 84.82$_{\color{blue}{-4.51}}$ & 83.48$_{\color{blue}{-0.55}}$ & 83.38$_{\color{blue}{-2.73}}$ & 72.42$_{\color{blue}{-4.02}}$ \\
\cdashline{1-5}[0.8pt/2pt]
{+ EnsAvg~\cite{kamnitsas2017ensembles}} & 90.21$_{\color{red}{+0.88}}$ & 88.39$_{\color{red}{+4.36}}$ & 88.77$_{\color{red}{+2.66}}$ & 80.73$_{\color{red}{+4.29}}$ \\
{+ EnsVot~\cite{kamnitsas2017ensembles}} & 90.12$_{\color{red}{+0.79}}$ & 87.18$_{\color{red}{+3.15}}$ & 88.09$_{\color{red}{+1.98}}$ & 79.61$_{\color{red}{+3.17}}$  \\
\cdashline{1-5}[0.8pt/2pt]
{+ All (TTA-EnsAvg)} & 84.16$_{\color{blue}{-5.17}}$ & 88.23$_{\color{red}{+4.20}}$ & 85.46$_{\color{blue}{-0.65}}$ & 75.52$_{\color{blue}{-0.92}}$ \\
{+ All (TTA-EnsVot)} & 84.16$_{\color{blue}{-5.17}}$ & 87.70$_{\color{red}{+3.67}}$ & 85.19$_{\color{blue}{-0.92}}$ & 75.05$_{\color{blue}{-1.39}}$ 
\end{tabular}
\label{tab10}}
\vspace{-4mm}
\end{center}
\end{table}

\subsubsection{Experimental Results}
Experiments are conducted on 2D-UNet~\cite{ronneberger2015u} and 3D-UNet~\cite{cciccek20163d}, and the results are presented in Table~\ref{tab9} and Table~\ref{tab10}. The results demonstrate that \textbf{these tricks can significantly improve overall performance for 2D-UNet~\cite{ronneberger2015u} on ISIC 2018~\cite{codella2019skin} and CoNIC~\cite{graham2021conic}, but not all of them are beneficial for 3D-UNet~\cite{cciccek20163d} on KiTS19~\cite{heller2021state} and LiTS~\cite{bilic2019liver}}. Among these tricks, TTA$_{\textrm{GTAug-B}}$ results in the maximum performance improvements, while TTA$_{\textrm{GTAug-A}}$ has the minimum performance improvements on average. Specifically, +EnsVot and +EnsAvg have small performance reductions on Precision by $-0.38\%$ and $-0.47\%$, respectively. Compared to results on ISIC 2018~\cite{codella2019skin} without using data augmentation, models using data augmentation show similar performance gains. Regarding the model ensemble on ISIC 2018~\cite{codella2019skin}, both EnsAvg~\cite{kamnitsas2017ensembles} and EnsVot~\cite{kamnitsas2017ensembles} improve model performance by $0.46\%$ Recall, $1.08\%$ Precision, $1.01\%$ Dice, and $0.92\%$ IoU, and $0.43\%$ Recall, $0.98\%$ Precision, $0.98\%$ Dice, and $0.89\%$ IoU, respectively.

Regarding the experimental results on 2D-UNet~\cite{ronneberger2015u} using the CoNIC~\cite{graham2021conic} dataset, \textbf{we observe improvements in all evaluation metrics, strongly demonstrating the impact of test time augmentation and model ensemble on medical image segmentation tasks~\cite{xu2022comprehensive,shorten2019survey,luo2022meta,han2022survey}}.However, compared to the experimental results of the baseline model on KiTS19~\cite{heller2021state} and LiTS~\cite{bilic2019liver}, results with TTA do not show significant improvements. The 3D setting model achieves the best performance of $93.00\%$, $98.73\%$, $94.39\%$, and $90.02\%$ on Recall, Precision, Dice, and IoU on setting-$\textrm{\romannumeral1}$ on KiTS19~\cite{heller2021state}, respectively. The 3D setting model also achieves the best performance of $27.65\%$, $56.65\%$, $29.44\%$, and $21.40\%$ on Recall, Precision, Dice, and IoU on setting-$\textrm{\romannumeral1}$ on LiTS~\cite{bilic2019liver}, respectively. From results on 3D-UNet~\cite{cciccek20163d} on KiTS19~\cite{heller2021state} and LiTS~\cite{bilic2019liver}, we can reach the same conclusions as in subsection~\ref{sec:3.3}.
\subsection{Result Post-Processing}
\label{sec:3.6}
Post-processing operations aim to enhance model performance via non-learnable approaches. For instance, segmentation results can be refined by integrating global image information. In this paper, we investigate two commonly used post-processing schemes in medical image analysis: all-but-largest-component-suppression (ABL-CS)~\cite{havaei2017brain} and removal of small areas (RSA)~\cite{chen2005hybrid}.
\begin{table*}[t]
\begin{center}
\renewcommand\arraystretch{1.5}
\setlength{\tabcolsep}{5pt}{
\caption{Experimental results on post-processing tricks. ``ABL-CS'' and ``RSA'' denotes all-but-largest-component-suppression strategy~\cite{dorent2022crossmoda} and remove small area strategy~\cite{chen2005hybrid}, respectively.}
\vspace{-3mm}
\begin{tabular}{ r | c c c c | c c c c} 
Methods & Recall (\%) & Percision (\%) & Dice (\%) & IoU (\%) & Recall (\%) & Percision (\%) & Dice (\%) & IoU (\%) \\ 
\hline \hline 
\multirow{2}{*}{2D-UNet~\cite{ronneberger2015u}} & \multicolumn{4}{c}{ISIC 2018~\cite{codella2019skin}} & \multicolumn{4}{c}{CoNIC~\cite{graham2021lizard}} \\
\cline{2-9}
~ & 88.18 & 89.88 & 86.89 & 85.80 & 78.12 & 77.25 & 77.23 & 77.58 \\
\cdashline{1-9}[0.8pt/2pt]
+ ABL-CS~\cite{saikumar2019deep} & 87.64$_{\color{blue}{-0.54}}$ & 90.57$_{\color{red}{+0.69}}$ & 86.99$_{\color{red}{+0.10}}$ & 86.00$_{\color{red}{+0.20}}$  & - & - & - & -\\
+ RSA~\cite{chen2005hybrid} & 88.15$_{\color{blue}{-0.03}}$ & 89.87$_{\color{blue}{-0.01}}$ & 86.91$_{\color{red}{+0.02}}$ & 85.83$_{\color{red}{+0.03}}$ & 78.00$_{\color{blue}{-0.12}}$ & 77.33$_{\color{red}{+0.08}}$ & 77.23$_{\color{red}{+0.00}}$ & 77.58$_{\color{red}{+0.00}}$ \\
\hline \hline
\multirow{2}{*}{3D-UNet~\cite{cciccek20163d}} & \multicolumn{4}{c}{KiTS19~\cite{heller2021state}:~settings-\romannumeral1} & \multicolumn{4}{c}{KiTS19~\cite{heller2021state}:~settings-\romannumeral2} \\
\cline{2-9}
~ & 91.01 & 95.20 & 92.50 & 87.35 & 27.35 & 46.71 & 29.63 & 21.51  \\
\cdashline{1-9}[0.8pt/2pt]
{+ ABL-CS~\cite{saikumar2019deep}} & 90.04$_{\color{blue}{-0.97}}$ & 95.63$_{\color{red}{+0.43}}$ & 92.14$_{\color{blue}{-0.36}}$ & 86.81$_{\color{blue}{-0.54}}$ & 23.49$_{\color{blue}{-3.86}}$ & 49.00$_{\color{red}{+2.29}}$ & 28.23$_{\color{blue}{-1.40}}$ & 21.26$_{\color{blue}{-0.25}}$  \\
{+ RSA~\cite{chen2005hybrid}} & 90.76$_{\color{blue}{-0.25}}$ & 95.33$_{\color{red}{+0.13}}$ & 92.41$_{\color{blue}{-0.09}}$ & 87.22$_{\color{blue}{-0.13}}$ & 26.82$_{\color{blue}{-0.53}}$ & 44.30$_{\color{blue}{-2.41}}$ & 29.20$_{\color{blue}{-0.43}}$ & 21.36$_{\color{blue}{-0.14}}$ \\
\end{tabular}
\label{tab11}}
\vspace{-5mm}
\end{center}
\end{table*}
\begin{table}[t]
\begin{center}
\renewcommand\arraystretch{1.5}
\setlength{\tabcolsep}{2.5pt}{
\caption{Experimental results on 3D LiTS dataset~\cite{bilic2019liver} on post-processing tricks. ``ABL-CS'' and ``RSA'' denotes all-but-largest-component-suppression strategy~\cite{dorent2022crossmoda} and remove small area strategy~\cite{chen2005hybrid}, respectively.}
\vspace{-3mm}
\begin{tabular}{ r | c c c c} 
 Methods & Recall (\%) & Percision (\%) & Dice (\%) & IoU (\%) \\ 
\hline \hline 
3D-UNet~\cite{cciccek20163d} & 89.33 & 84.03 & 86.11 & 76.44 \\ 
\cdashline{1-5}[0.8pt/2pt]
{+ ABL-CS~\cite{saikumar2019deep}} & 89.31$_{\color{blue}{-0.02}}$ & 87.38$_{\color{red}{+3.35}}$ & 87.79$_{\color{red}{+1.68}}$ & 79.13$_{\color{red}{+2.69}}$ \\
{+ RSA~\cite{chen2005hybrid}} & 89.32$_{\color{blue}{-0.01}}$ & 84.19$_{\color{red}{+0.16}}$ & 86.19$_{\color{red}{+0.08}}$ & 76.58$_{\color{red}{+0.14}}$ 
\end{tabular}
\label{tab12}}
\vspace{-4mm}
\end{center}
\end{table}

\subsubsection{ABL-CS} ABL-CS~\cite{havaei2017brain} aims to remove some wrong areas in the segmentation results based on knowledge of the organism's physical properties~\cite{dorent2022crossmoda}. For example, for the heart segmentation task, we all know that every person has only one heart, so if there are small segmentation areas in the obtained mask, we need to remove this small areas~\cite {saikumar2019deep}. 
 
\subsubsection{RSA} For MedISeg, the imaging protocol is generally unchanged, such that the area of each instance segmentation mask remains unvaried as well. Based on this physical property, we can set a pixel-level threshold to remove some instance masks that are too small (\ie, under the given threshold) in the obtained segmentation masks~\cite{chen2005hybrid}. In our work, following~\cite{mahmood2018unsupervised,sutherland2019applying}, the threshold is set to $5000$ for setting-$\textrm{\romannumeral1}$, $80$ for for setting-$\textrm{\romannumeral2}$, $120$ for ISIC 2018~\cite{codella2019skin}, and $10$ for CoNIC~\cite{graham2021conic} and LiTS~\cite{bilic2019liver}. This means that all masks in the segmented area with less than this threshold are eliminated from the final results.

\subsubsection{Experimental Results}
Experiments for post-processing strategies are carried out on 2D-UNet~\cite{ronneberger2015u} for ISIC 2018~\cite{codella2019skin} and CoNIC~\cite{graham2021conic} datasets, and 3D-UNet~\cite{cciccek20163d} for KiTS19~\cite{kutikov2009renal} and LiTS~\cite{bilic2019liver} datasets, respectively. Results are shown in Table~\ref{tab11} and Table~\ref{tab12}. We can observe that compared to the baseline 2D-UNet on ISIC 2018~\cite{codella2019skin}, implementing ABL-CS on the baseline model can boost three-quarters of the evaluation metrics. For example, $0.69\%$ Precision, $0.10\%$ Dice, and $0.20\%$ IoU. Implementing RSA on the baseline model can boost $0.02\%$ Dice, and $0.03\%$ IoU. We can also observe that +RSA has a weak effect on performance improvements on ISIC 2018~\cite{codella2019skin}, \eg, the increased model performance is almost always less than 0.1\%.
Since ABL-CS does not apply to the CoNIC~\cite{graham2021conic} dataset, we only implement RSA on CoNIC~\cite{graham2021conic}. We can observe that values under these evaluation metrics have very little changes.
The experimental results on 3D-UNet~\cite{cciccek20163d} on KiTS19~\cite{kutikov2009renal} show that + ABL-CS can improve model performance on Precision by $0.43\%$ and $2.29\%$ on setting-$\textrm{\romannumeral1}$ and setting-$\textrm{\romannumeral2}$, respectively. Besides, + RSA only works on setting-$\textrm{\romannumeral1}$ by improving $0.13\%$ Precision. 
\textbf{The above experimental results of + ABL-CS and + RSA on the 3D setting on KiTS19~\cite{kutikov2009renal} demonstrate that these two post-processing schemes are ineffective in improving performance for Recall, Dice, and IoU.}
The experimental results on LiTS~\cite{bilic2019liver} show that these two schemes can improve all evaluation metrics except Recall. These experimental results show that the performance of the same post-processing operation on different datasets and different evaluation metrics is also different. It also shows us that we need to pay attention to the state of the datasets when choosing the post-processing tricks, and to what evaluation metrics we are focusing on.
\section{Discussions}
\label{sec:4}
In this section, we discuss the potential challenges and problems corresponding to the above experimental tricks. In addition, we also point out the limitations of these tricks. As illustrated in Figure~\ref{fig2}, an unabridged MedISeg system can be separated into six main implementation phases. We collected and explored a series of tricks in each phase. The potential challenges we addressed included but not limited to \uline{\textbf{small dataset}}, \uline{\textbf{class imbalance learning}}, \uline{\textbf{multi-modality learning}}, and \uline{\textbf{domain adaptation}}.

\begin{figure*}[t]
\centering
\includegraphics[width=.99\textwidth]{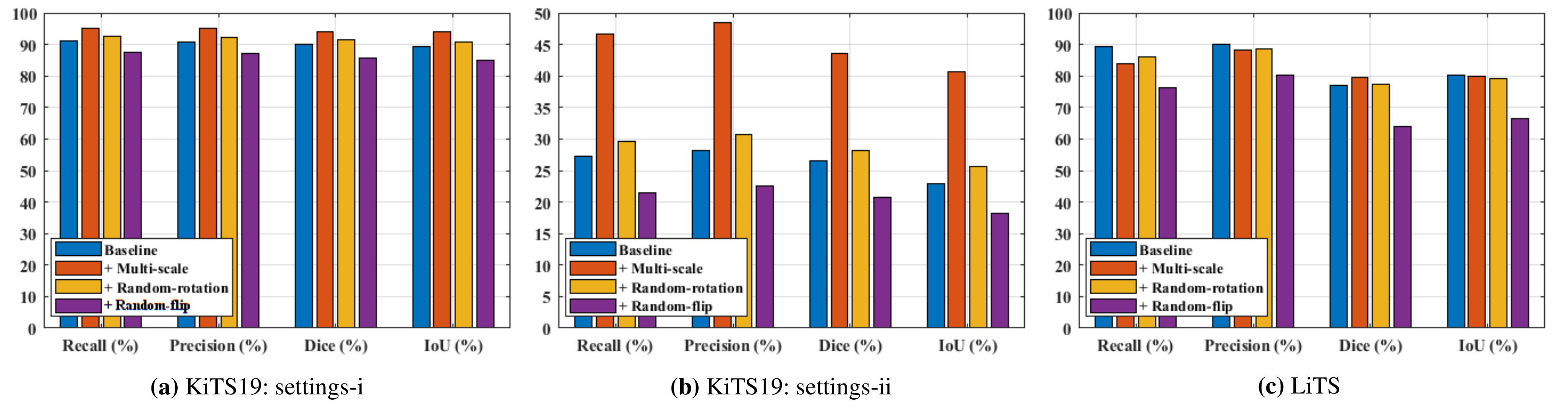}
\vspace{-2mm}
\caption{The segmentation result illustrations by adding each single data augmentation scheme on 3D KiTS19~\cite{heller2021state} and LiTS~\cite{bilic2019liver} datasets, respectively. The baseline model is 3D-UNet~\cite{cciccek20163d}. (a) the foreground ``kidney'' and the ``background'' in KiTS19~\cite{heller2021state}, (b) the foreground ``tumor'' and the ``background'' in KiTS19~\cite{heller2021state}, (c) the foreground ``liver'' and the ``background'' in LiTS~\cite{bilic2019liver}.}
\label{fig4}
\end{figure*} 
\subsection{Challenges} 
By incorporating the collected tricks into various baseline models, we observed significant improvements in performance. It is worth noting that while these enhancements can be attributed to the implementation of the tricks on the baselines, the performance gains are also attributable to the underlying fundamental challenges posed by medical image processing problems themselves. Despite differences in deployment and appearance, the problems addressed by the tricks share similar objectives with advanced approaches. For instance, we observed that fine-tuning pre-trained weights can lead to substantial enhancements in model performance (\textit{ref.}~Table~\ref{tab1} and Table~\ref{tab2}).
First, this observation demonstrates that the pre-trained weights on the natural scenes can be transferred into the medical scene via a plain fine-tuning strategy, thereby essentially solves \uline{\textbf{multi-modality learning}} and \uline{\textbf{domain adaptation}} (\ie, cross-domain representations between natural scenarios and medical scenarios) problems. Although these problems can also be solved by some advanced transfer learning~\cite{pan2009survey,weiss2016survey,tan2018survey} or meta-learning approaches~\cite{hospedales2020meta,luo2022meta}, fine-tuning is a cheap solution. Second, it shows that training on more data samples helps to improve the model performance, which corresponds to problems of \uline{\textbf{small dataset}} and overfitting. These two problems are accentuated in models with very strong learning capabilities (\eg, the transformer-based models), as such models generally require very large datasets to train.

In addition to the pre-training model, observations and conclusions can also be made on data pre-processing, data augmentation, and test time augmentation, as shown in Table~\ref{tab3} to Table~\ref{tab12}. Data pre-processing and data augmentation methods can enhance the experimental dataset representation ability, implicitly addressing challenges posed by insufficient training samples, \textbf{small dataset}, and \textbf{class imbalance learning}. Patching, OverSam, and ReSam methods are effective in mitigating these challenges. In particular, Patching and OverSam can be used to address \textbf{small dataset}, while Patching, OverSam, ReSam, and IntesNorm can be employed to address \textbf{class imbalance learning}. Fortunately, these methods do not incur significant computational costs and can be implemented seamlessly. While data augmentation and test time augmentation may not always improve performance, this may be due to dataset-specific factors rather than limitations of the techniques themselves.

In addition, there are significant differences between individual geometric transformation-based data augmentation schemes. Segmentation illustrations of each data augmentation scheme on KiTS19~\cite{kutikov2009renal} and LiTS~\cite{bilic2019liver} are visualized in Figure~\ref{fig4}. Multiple scales and rotation improve model performance on some evaluation metrics for KiTS19~\cite{kutikov2009renal}, while the use of flip markedly reduces the model performance. Results on LiTS~\cite{bilic2019liver} show that multi-scale only improves a portion of the evaluation metrics (\eg, Dice), and the remaining data augmentation schemes do not work. These results validate our hypothesis that data augmentation methods with non-destructive inductive biases (\eg, multi-scale) are more beneficial for images with relatively fixed object positions. GAN-based data augmentation methods can improve dataset representation without destroying such inductive biases, achieving better performance on KiTS19~\cite{kutikov2009renal} (see Table~\ref{tab5}). These experiments suggest that future data augmentation designs should consider the specific state (\eg, modality, distribution, and lesion size) of the dataset used. Additionally, \textbf{class imbalance learning} can also be addressed using the collected tricks (see Table~\ref{tab7} and Table~\ref{tab8}). In summary, our work provides not only a completed MedISeg survey but also a practical guide for addressing segmentation-related challenges in medical image processing, including but not limited to \textbf{small dataset}, \textbf{class imbalance learning}, \textbf{multi-modality learning}, and \textbf{domain adaptation}.

\subsection{Limitations} 
There are some limitations of this work. For example, the collected tricks are limited, meaning that the tricks collected and experimented with in our paper are only a finite subset of all the medical image segmentation tricks. This point can be elaborated into the following two aspects:

\myparagraph{(1)~Surveyed Subset of Tricks:} We only collected a representative number of tricks for the same category of phase. For example, there exist many data augmentation schemes such as translating, color jittering, contrast, simulation of low resolution, Gaussian noise injection, mixing images, random erasing, Gaussian blur, mixup, and cutmix~\cite{shorten2019survey,xu2022comprehensive,yeung2022unified,hataya2022meta,zhang2020feature}. However, due to the practical requirements of medical image processing and the characteristics of the experimental datasets, only a small number of these techniques were used.

\myparagraph{(2)~Limited Application of Surveyed Tricks:} In our work, we chose commonly used 2D-UNet~\cite{ronneberger2015u} and 3D-UNet~\cite{cciccek20163d} as our baseline models and validated the effectiveness of these tricks on four representative 2D and 3D medical image datasets. However, there are a large number of progressive MedISeg models with different backbones at different levels that were not involved.

\myparagraph{(3)~Limited Effect of Surveyed Tricks:} Medical image processing is closely related to clinical practice and involves many complex challenging problems, such as sheltered objects~\cite{zhou2019design}, blurred images~\cite{na2018guided}, the long-tail problem~\cite{roy2022does}, incremental learning problem in medical images~\cite{kumar2018example}, and object boundary detection and refinement~\cite{wang2022boundary}. In our work, we only addressed some of the fundamental challenges.

Overall, our work provides a useful starting point for exploring a subset of medical image segmentation tricks, but further investigation is needed to fully explore the potential of these techniques in medical image processing. 
\section{Conclusions and Future Directions}
\label{sec:5}
In this work, we have collected a comprehensive set of MedISeg tricks that cover common and fundamental schemes used in medical image segmentation. To avoid performance ambiguity from implementation variations, we experimentally explored the effectiveness of these tricks on consistent 2D-UNet~\cite{ronneberger2015u} and 3D-UNet~\cite{cciccek20163d} baseline models. Our experiments on 2D ISIC 2018~\cite{codella2019skin}, 2D CoNIC~\cite{graham2021lizard,graham2021conic,graham2019hover}, 3D KiTS19~\cite{heller2021state} and 3D LiTS~\cite{bilic2019liver} datasets have shown the effectiveness of these tricks and helped us provide empirical guidance for the forthcoming segmentation pipelines, including network architectures, training strategies, and loss functions.

One of the important contributions of our work is the explicit exploration of the effect of these collected tricks. Our work not only promotes subsequent methods to pay attention to tricks but also achieves a fair result comparison. This is especially necessary as network architectures become more sophisticated in the face of complex tasks such as image segmentation~\cite{zhang2021self,zhang2018context}, object detection~\cite{lin2017feature,lin2017focal}, and image generation~\cite{zhou2019review,shin2018medical}. Compared to existing paper-driven segmentation surveys~\cite{xu2022comprehensive,shorten2019survey,han2022survey,mainak2019state,hesamian2019deep,yi2019generative,zhou2019review,munir2019cancer} that only blandly focus on advantage and limitation analyses, our work provides extensive experiments and is more technically operable.

In the future, we plan to work in the following directions:

\subsection{Trick Track}

\myparagraph{(1)~Survey and develop more tricks on MedISeg.} As MedISeg is closely integrated with clinical practice, it is of great practical value to continue exploring and developing more advanced MedISeg tricks to meet the requirements of different problems.

\myparagraph{(2)~Explore the effectiveness of tricks on more methods and datasets.} To enable a more comprehensive and fair comparison of experimental results, it is necessary to conduct a thorough trick survey. This is especially important when faced with MedISeg problems where different image types, distributions, and inner class divergence can affect the effectiveness of a specific trick.

\myparagraph{(3)~Explore trick-inspired model designs.} Although tricks are often overlooked in existing publications, the principles and ideas they contain can inspire subsequent work to achieve cheaper and more computationally friendly model designs.

\myparagraph{(4)~Explore attention-based tricks.} With the multi-head attention mechanism providing a strong feature representation ability, the visual transformer framework has received more attention in communities of computer vision and medical image analysis. However, due to its complex internal architectures and immature practice applications (especially in the face of small datasets), there is a need to explore trick research on the vision transformer framework.

\subsection{After Large Vision Models}
Recently, there has been a surge of interest in large vision models, such as SAM~\cite{kirillov2023segment} and MedSAM~\cite{ma2023segment}, which have ushered in a new era of image semantic recognition, particularly for dense image predictions, such as semantic segmentation and instance segmentation~\cite{yu2023inpaint,he2023accuracy}. These models have enabled breakthrough progress and impressive performance on challenging tasks of images of natural scenes, such as image inpainting and camouflaged object detection~\cite{tang2023can,deng2023segment,he2023accuracy,zhou2023can,yu2023inpaint,hu2023sam}. However, there still exists a gap between the performance of large vision models on specific medical datasets and that of state-of-the-art models~\cite{huang2023segment,wu2023medical,cheng2023sam,zhang2023customized}. Therefore, it is imperative to explore integrate off-the-shelf large vision models with downstream medical image processing tasks and achieve satisfactory results on specific medical image datasets. Moreover, there is great potential in combining large vision models with large vision-language models (\eg, ChatCAD~\cite{wang2023chatcad} and CLIP~\cite{radford2021learning}) in the domain of medical image analysis. This could involve designing efficient visual prompts~\cite{zhou2022learning} or visual adapters~\cite{zhang2023adding} that would enable the resulting models to be deployed on edge computing devices. This promising research direction could lead to significant advancements in MedISeg and other medical image analysis tasks.
\normalem
\bibliographystyle{IEEEtran}
\bibliography{main}
\end{document}